\documentclass[10pt,twocolumn,letterpaper]{article}

\PassOptionsToPackage{table}{xcolor}

\usepackage[pagenumbers]{cvpr} 

\usepackage{amsthm,mathtools}
\usepackage{microtype}
\usepackage{multirow}
\usepackage{algorithm}
\usepackage{algpseudocode}
\usepackage{listings}

\DeclareMathOperator*{\argmin}{arg\,min}

\definecolor{cvprblue}{rgb}{0.21,0.49,0.74}
\usepackage[hypertexnames=false,pagebackref,breaklinks,colorlinks,allcolors=cvprblue]{hyperref}

\def\confName{CVPR}
\def\confYear{2027}

\title{RefMix: Expanding Endpoint Proposals for Diversity-Preserving Reward Fine-Tuning}

\author{
Jaemin Kim$^{1,2,\dagger}$ \quad
Christian Simon$^{2}$ \quad
Dongseok Shim$^{2}$ \quad
Wei-Yao Wang$^{2}$ \quad
Qiyu Wu$^{2}$ \\
Masato Ishii$^{2}$ \quad
Shusuke Takahashi$^{2}$ \quad
Jong Chul Ye$^{1,*}$ \quad
Yuki Mitsufuji$^{2,*}$ \\
$^{1}$KAIST \qquad $^{2}$Sony Group Corporation 
\\
{\small $^{*}$Corresponding authors.}
}

\newcommand{\sg}{\operatorname{sg}}
\newcommand{\ESS}{\operatorname{ESS}}
\newcommand{\DP}{\boldsymbol{D}_{\mathrm{all}}}
\newcommand{\DQ}{\boldsymbol{D}_{\mathrm{rew}}}
\newcommand{\DLPIPS}{\boldsymbol{D}_{\mathrm{LPIPS}}}
\providecommand{\citep}[1]{\cite{#1}}
\newtheorem{proposition}{Proposition}

\definecolor{addbg}{RGB}{232,246,232}
\definecolor{addfg}{RGB}{0,110,0}

\def\Vec#1{{\boldsymbol{#1}}}

\begin{document}

\twocolumn[{%
\renewcommand\twocolumn[1][]{#1}%
\maketitle
\vspace{-0.8cm}
\begin{center}
\centering
\captionsetup{type=figure}
\includegraphics[width=\textwidth]{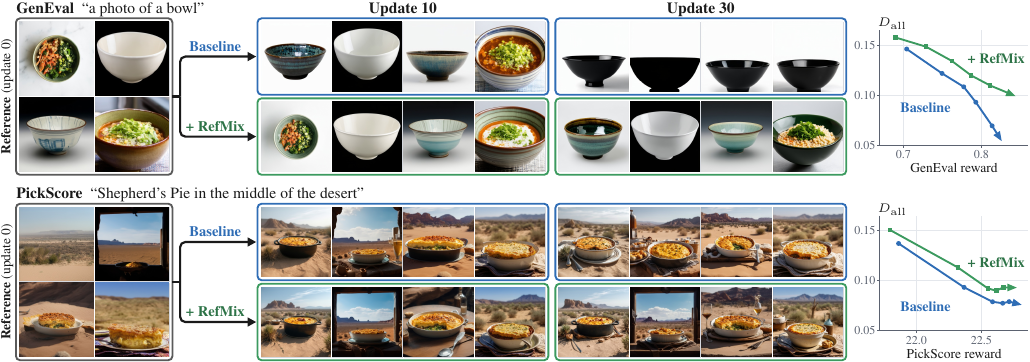}
\vspace{-0.6cm}
\captionof{figure}{Reward fine-tuning with RAM, without and with RefMix. Each column shares one initial noise across the reference model (update 0) and updates 10 and 30. Baseline samples converge to similar layouts and appearances, whereas RefMix samples remain varied. Right: $D_{\mathrm{all}}$ (DINO diversity) against held-out reward over updates 5 to 30; RefMix lies above the baseline at matched reward.}
\label{fig:qualitative-comparison}
\end{center}%
\vspace{0.2cm}
}]

\begin{abstract}
\vspace{-1cm}

Online forward matching aligns flow-based generative models by repeatedly generating, rewarding, and fitting model-produced endpoints. However, this behavior creates an on-policy feedback loop: as high-reward modes become increasingly favored, alternative modes become less likely to appear in subsequent training pools, progressively reducing visual diversity. We introduce RefMix, a reference-assisted endpoint proposal strategy that addresses this limitation by injecting alternative candidates into the training pool while retaining the original task reward and forward matching objective. In particular, RefMix uses a reference model during early high-noise steps to expose alternative visual modes, followed by the current reward-tuned model for endpoint completion. Across state-of-the-art online forward matching methods, RefMix consistently improves visual diversity while preserving task reward performance. Controlled inference experiments further demonstrate that training-time endpoint exposure is critical to improve reward--diversity trade-offs, beyond the impact of reference-assisted inference. By rethinking how training endpoints are proposed, RefMix achieves up to 44\% higher diversity over forward matching baselines at comparable reward levels.
Project page: \url{https://kjm981995.github.io/RefMix}
{\let\thefootnote\relax\footnotetext{$^{\dagger}$Work done during an internship at Sony Group Corporation.}}
\end{abstract}

\section{Introduction}

Reward optimization can improve alignment while progressively reducing the diversity of what a generative model produces. High-reward modes become increasingly dominant, while alternative visual outputs gradually disappear from the model distribution ~\citep{barcelo2024hrf,he2025gardo,liu2026drift,tan2026perceptualentropy}, as in the baseline rows of Figure~\ref{fig:qualitative-comparison}. This issue creates a fundamental challenge for online forward matching~\citep{xue2025awm,zheng2025diffusionnft,bergmeister2026ram}, which repeatedly uses reward-scored samples from the current model as endpoints for subsequent updates. Once diverse endpoints become underrepresented in the current model, they also become scarce among future training targets, creating a feedback loop that further concentrates the model distribution. 

Once the current model's rollouts concentrate around a small set of samples, reweighting the observed endpoints cannot provide direct supervision from alternatives absent from the endpoint pool. Existing approaches counter this contraction by adding perceptual feedback to the reward signal \citep{liu2026drift,tan2026perceptualentropy}, or by selecting candidates from an enlarged on-policy pool \citep{liu2026drift}. These approaches change how current-model candidates are scored. However, their direct endpoint supervision still depends on the candidates that the current sampler produces.

This suggests that the diversity bottleneck might arise before reward scoring and sample selection, at the stage where training endpoints are obtained. If the endpoint pool already covers only a restricted set of visual modes, subsequent reweighting or selection can only operate within that restricted support. Thus, an important yet underexplored question is whether diversity can be preserved by reconsidering how endpoints enter the training pool, rather than  modifying how they are rewarded.

Motivated by the observed diversity problems, we introduce RefMix, a reference-assisted endpoint proposal strategy that maintains access to diverse training targets throughout alignment while leaving the reward objective and forward matching formulation unmodified. An offline reference bank offers a simple alternative to purely on-policy sampling, but it becomes increasingly mismatched as the reward-tuned model evolves. RefMix instead generates endpoints online by using the reference model during early high-noise steps to introduce alternative visual modes, followed by the current model to complete them with the reward-adapted behavior. This strategy produces diverse yet adaptive training targets that evolve alongside the model throughout alignment.

RefMix improves diversity during reward alignment without modifying the task reward, forward matching objective, or denoising budget. Across state-of-the-art online forward matching methods, including NFT~\citep{zheng2025diffusionnft}, AWM~\citep{xue2025awm}, and RAM~\citep{bergmeister2026ram}, RefMix preserves higher visual diversity while maintaining competitive reward performance across multiple alignment tasks. On PickScore~\citep{kirstain2023pick}, for instance, reward remains within $\sim$0.3\% of the original methods. On OCR, every method gains diversity at equal or higher reward, and on GenEval~\citep{ghosh2023geneval}, RAM gains $71\%$ diversity while its reward rises from $0.822$ to $0.837$, demonstrating that the benefits generalize across different alignment objectives and forward matching methods. 

During inference, applying the same reference prefix to forward matching baselines (\eg, NFT~\cite{zheng2025diffusionnft}) can also increase diversity, albeit at lower reward levels. As an example, under the same reward level, our RefMix improves up to $\sim$44.2\% more diversity of generated samples compared to the NFT baseline. 
This improvement demonstrates that the benefit extends beyond hybrid sampling and stems from exposing the model to diverse endpoints during training.
To summarize, the contributions of our proposed work are:

\begin{enumerate}
\item We identify candidate exposure as an overlooked bottleneck and establish endpoint proposal as a novel strategy for online forward matching.
\item We introduce \textbf{RefMix}, an online hybrid proposal with a reference prefix and a current-model suffix, preserving the pool size and number of denoising evaluations. Offline-bank controls test the importance of generating auxiliary endpoints with the evolving model.
\item Extensive experiments establish the effectiveness and generality of RefMix across diverse alignment tasks and state-of-the-art forward matching methods. Evaluations span multiple benchmarks, including PickScore, GenEval, and OCR, together with controlled inference interventions and proposal ablations that reveal how training endpoint composition shapes the reward-diversity trade-off.

\end{enumerate}

\section{Related Work}
\label{sec:related-work}

\paragraph{Online reward fine-tuning for diffusion and flow models.}
Online reward fine-tuning repeatedly generates samples, evaluates their rewards, and updates the model; two common approaches differ in how they use this feedback. Trajectory-based methods formulate reverse denoising as a stochastic policy and apply policy gradients \citep{liu2025flowgrpo}. Forward matching methods instead retain the supervised noising-and-regression structure, while incorporating reward through sample weighting or target modification \citep{zheng2025diffusionnft,xue2025awm,kveton2026advantageflow,bergmeister2026ram,fu2026flowawr}. 
These forward matching updates can be related through an anchored reward-weighted regression view, with method-specific coefficients and anchor fields \citep{lee2026rsm,choi2026scaling}. These methods determine how sampled endpoints affect the model, but the direct reward signal remains limited to endpoints in the generated pool. RefMix instead changes how the endpoints that provide this supervision are generated, while retaining the method-specific regression form.

\paragraph{Diversity-preserving alignment.}
Prior approaches to preserving diversity include explicit perceptual objectives, reference regularization, and candidate selection. Perceptual-entropy regularization \citep{tan2026perceptualentropy} directly constrains the spread of generated samples in a representation space. GARDO \citep{he2025gardo} applies adaptive reference regularization and amplifies rewards for diverse high-quality samples. DRIFT \citep{liu2026drift} selects a perceptually compact subset from an enlarged on-policy pool to suppress reward outliers that can accelerate collapse. These approaches intervene in the training objective or in the selection of generated candidates. Perceptual objectives and reference penalties require balancing an additional training term against task reward; enlarged-pool selection requires extra generations and selects only among current endpoints. RefMix instead broadens a fixed-size training pool without adding a regularization term or extra generations.

\paragraph{Model composition across parameters and denoising time.}
Model composition offers a way to combine reference and reward-tuned behavior when generating training endpoints, either through model parameters or across denoising time. In parameter space, previous methods \citep{rame2023rewardedsoups, biggs2024diffusionsoup} interpolate policies specialized to different rewards or data shards. Sampling-time approaches instead use different models over different denoising intervals. ImageReFL \citep{sorokin2025imagerefl} uses the base model at high noise and a reward-tuned model at low noise to control an inference-time quality--diversity trade-off, not using endpoints from combined generation as an online forward matching pool. Diffusion Blend \citep{cheng2026diffusion} blends the backward processes of several reward-tuned models at inference time, selecting a reward combination or regularization strength without retraining. P-GRAFT \citep{anil2026pgraft} uses reference-model rollouts to evaluate intermediate states, trains the high-noise segment toward selected intermediate states, and retains the reference model for the remaining low-noise segment. RefMix uses temporal composition to generate clean endpoints for subsequent online forward matching updates.

\section{Background and Motivation}
\label{sec:diagnosis}

\subsection{Forward matching for reward tuning}
\label{sec:forward-reward-setup}

Let $\Vec{v}_\theta$ denote the trainable velocity field and $\Vec{v}_{\mathrm{old},n}$ the rollout velocity field held fixed during online update $n$. Let $P_n(\cdot\mid\Vec{c})$ denote the current endpoint distribution generated by sampling with $\Vec{v}_{\mathrm{old},n}$ for prompt $\Vec{c}$, and let $\{\Vec{x}_0^i\}_{i=1}^N\sim P_n(\cdot\mid\Vec{c})$ denote its candidate pool.
Forward matching reward-tuning methods generate this pool, score its endpoints, and compute a method-specific reward signal $\Vec{A}_i$ within each prompt group:
\begin{equation}
\begin{gathered}
    {r}_i=\texttt{Reward}(\Vec{x}_0^i, \Vec{c}), \qquad
  \Vec{A}_i=\mathcal{A}\!\left({r}_i,\{{r}_j\}_{j=1}^{N}\right),
\end{gathered}
\label{eq:on-policy-template}
\end{equation}
where $\mathcal{A}$ denotes a normalizing advantage rule. Each clean endpoint is then noised through the forward path used in conditional flow matching:
\begin{equation}
\begin{gathered}
  \Vec{x}_t^i=a_t \Vec{x}_0^i+b_t\Vec{\epsilon}_i,\qquad \Vec{u}_t^i=\dot a_t \Vec{x}_0^i+\dot b_t\Vec{\epsilon}_i. 
\end{gathered}
  \label{eq:generic-analytic-noising}
\end{equation}
Here,  $\Vec{\epsilon}_i$ denotes noise sampled independently of $\Vec{x}_0^i$, and $a_t,b_t$ specify
the forward path. The noisy state $\Vec{x}_t^i$ and conditional velocity $\Vec{u}_t^i$ form a supervised forward matching pair. For brevity, we omit shared time and prompt arguments; \eg, $\Vec{v}_{\Vec{\theta}}(\Vec{x}_t^i)$ denotes $\Vec{v}_{\Vec{\theta}}(\Vec{x}_t^i,t,\Vec{c})$. 

Following the reward-based velocity-matching view in~\citep{choi2026scaling}, we relate the velocity updates of the methods considered here to an anchored regression $\mathcal{L}_\mathrm{FM}^i(\Vec{\theta})$ of the form as follows: 
\begin{equation}
\begin{aligned}
\mathcal{L}_\mathrm{FM}^i(\theta) &= \Vec{\lambda}_i(t,\Vec{A}_i)\left\|\Vec{v}_\Vec{\theta}(\Vec{x}_t^i)-\Vec{u}_t^i
\right\|^2 \\
& + \Vec{g}_i(t,\Vec{A}_i)\left\|\Vec{v}_\Vec{\theta}(\Vec{x}_t^i)-\Vec{v}_{\texttt{anc}}
(\Vec{x}_t^i) \right\|^2,
\end{aligned}
\label{eq:anchored-forward-regression}
\end{equation}
where $\Vec{v}_{\texttt{anc}}$ is the method-specific anchor field, and $\Vec{\lambda}_i(t,\Vec{A}_i)$ and $\Vec{g}_i(t,\Vec{A}_i)$ are coefficients on the endpoint and anchor terms, respectively. We provide the method-specific forms in App.~\ref{app:forward-objectives}, including the loss identity for NFT~\citep{zheng2025diffusionnft} and the corresponding gradient forms for AWM~\citep{xue2025awm} and RAM~\citep{bergmeister2026ram}.

Every term in Eq.~\eqref{eq:anchored-forward-regression} is evaluated over the sampled pool, such that endpoints outside the pool provide no direct supervision. Because the pool is drawn from the currently evolving model distribution $P_n$, each update depends on the model's own generations, creating a feedback loop that can progressively concentrate the output distribution.

\subsection{Online sampling limits endpoint diversity}
\label{sec:online-sampling}
\begin{figure}[t]
\centering
\includegraphics[width=0.88\linewidth]{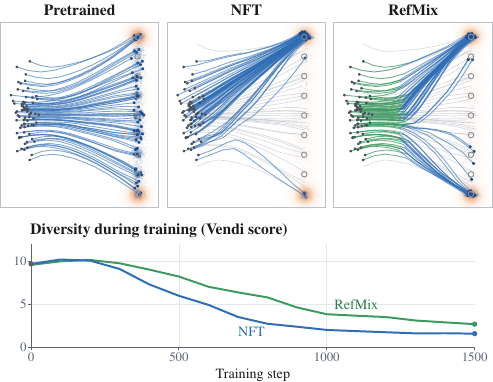}
\vspace{-0.2cm}
\caption{On-policy contraction in 2D toy flow with nine data modes, of which the two end modes (orange) have high reward. Top: sampling trajectories of the pretrained model, NFT, and RefMix; green segments are the reference prefix. NFT sends most trajectories to one of the two high-reward modes, whereas RefMix keeps both. Bottom: Vendi score \citep{friedman2022vendi} over training.}
\label{fig:toy-contraction}

\end{figure}

As follows, we examine how the feedback loop can progressively reduce output diversity. 
Consider a distinct and high-reward region $S$ with probability $p_n(S\mid\Vec{c})$ under the current model. For a pool of $N$ independently sampled endpoints, the probability that at least one endpoint from $S$ appears in the pool is:
\begin{equation} 
P(S) = 1-\left(1-
p_n(S\mid\Vec{c})\right)^N 
\approx N p_n(S\mid\Vec{c}), 
\end{equation} 
where the approximation holds when $Np_n(S\mid\Vec{c})\ll1$. When $S$ is absent from the pool, it receives no direct supervision and consequently appears less frequently in subsequent endpoint pools. The direct supervision therefore diminishes, reinforcing the contraction of the endpoint distribution across updates. Figure~\ref{fig:toy-contraction} illustrates this contraction in a toy example (see details in App.~\ref{app:amplification-toy}), where the diversity of NFT decreases steadily over training. Increasing $N$ can partially improve coverage, but incurs additional generation cost and remains limited by the current model's coverage. Thus, this feedback loop suggests that preserving diversity requires considering not only how endpoints are rewarded, but also which endpoints are exposed to each update.

 \begin{figure*}[t]
\centering
\includegraphics[width=1\textwidth]{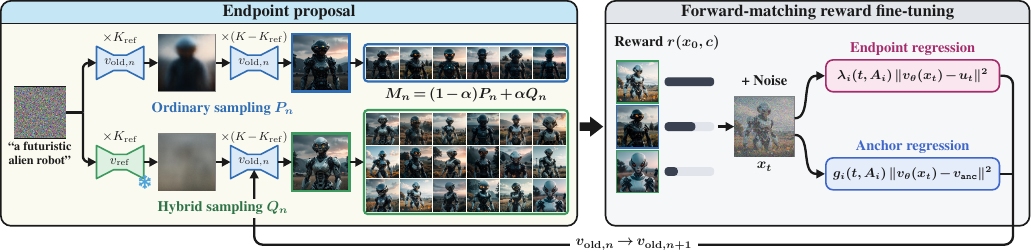}
\caption{RefMix replaces the first $K_{\mathrm{ref}}$ of $K$ denoising calls with the frozen reference field to form the hybrid proposal $Q_n$, and mixes them with ordinary endpoints from $P_n$ into the training pool drawn from $M_n$. The pool is scored, re-noised, and fitted with the endpoint and anchor regressions of Eq.~(3) before the rollout field is updated.}
\label{fig:refmix-pipeline} 
\end{figure*}

\section{Reference-Assisted Endpoint Proposals}
\label{sec:method}

Section~\ref{sec:diagnosis} shows that the update depends on which endpoints are exposed. This section asks how an auxiliary proposal can add such exposure while remaining compatible with the current model. An offline reference bank adds exposure but drifts from the current model, which motivates RefMix as an online proposal that follows it.

\subsection{Mixing endpoint proposals}
\label{sec:proposal-mixing}

Let $Q_n(\cdot\mid \Vec{c})$ be an auxiliary endpoint proposal at update $n$. For a given prompt $\Vec{c}$, we combine it with the current-model proposal $P_n$ through
\begin{equation}
M_n=(1-\alpha){P}_n+\alpha {Q}_n, 
\label{eq:proposal-mixture}
\end{equation}
where $\alpha\in[0,1]$ is the fraction of auxiliary endpoints. For a pool of $N$ endpoints, $\alpha N$ are supplied by $Q_n$ and the rest by $P_n$. All endpoints receive the original task reward with advantages computed within the prompt group by Eq.~\eqref{eq:on-policy-template}.

The resulting endpoints are re-noised using Eq.~\eqref{eq:generic-analytic-noising} and passed to the method-specific regression in Eq.~\eqref{eq:anchored-forward-regression}. Thus, the reward function, coefficient rules, and anchor are retained, while the endpoint proposal changes. Since advantages depend on the endpoints in the group, their values can change with the proposal even when the advantage rule is unchanged. Setting $\alpha=0$ recovers the ordinary current-model pool. We illustrate the mixing endpoint proposals in Figure~\ref{fig:refmix-pipeline}.

\subsection{Candidate exposure and the forward update}
\label{sec:proposal-analysis}

An auxiliary proposal can increase the exposure of a high-reward visual mode that is underrepresented in $P_n$, provided that $Q_n$ assigns sufficient probability mass to the mode. Thus, the benefit comes from the complementary coverage provided by $Q_n$, rather than from mixing itself. 

A direct choice of $Q_n$ is an offline reference bank consisting of endpoints generated in advance for each training prompt and reused during training (see App.~\ref{app:experimental-details}).
Since fine-tuning starts from the reference model, a mode whose probability has decreased during training is still generated more often by the reference model than by the current model. A reference bank therefore increases the exposure of such a mode. However, exposure alone does not determine how the exposed endpoints affect the model update, which depends on their regression targets and advantages in Eq.~\eqref{eq:anchored-forward-regression}.

\paragraph{Why the auxiliary proposal should follow the current model.}

We examine how an auxiliary endpoint contributes to the forward matching update. 
For NFT, the influence of an auxiliary endpoint can be characterized explicitly through the pointwise population regression solution. For a fixed rollout field and branch coefficient $\beta>0$, the solution at $(\Vec{x}_t,t,\Vec{c})$ satisfies:
\begin{equation}
\Vec{v}_n^\star-\Vec{v}_{\mathrm{old},n}
=\frac{1}{\beta}\mathbb E_{M_n}\!\left[
\Vec{A}_i\bigl(\Vec{u}_t-\Vec{v}_{\mathrm{old},n}\bigr)\mid
\Vec{x}_t,t,\Vec{c}\right].
\label{eq:main-nft-proposal-update}
\end{equation}

Changing the proposal therefore affects not only which endpoints are observed, but also the group-dependent weights within the expectation. Each endpoint contributes a reward-weighted residual between the target velocity $\Vec{u}_t$ and the rollout-model prediction $\Vec{v}_{\mathrm{old},n}$. These two components, the advantage weight $\Vec{A}_i$ and the residual, determine whether an auxiliary endpoint provides effective supervision. 
Since auxiliary and current-model endpoints share a prompt group, each advantage is computed relative to both sources. If rewards from auxiliary endpoints are consistently lower than those of current-model endpoints, most auxiliary endpoints receive negative advantages. In that case, the exposure added by $Q_n$ no longer provides effective supervision, so rewards from auxiliary endpoints should remain comparable to those of $P_n$.

Similarly, the residual grows as an endpoint moves away from $P_n$, and endpoints with a large difference dominate the regression. Auxiliary endpoints should therefore remain close to $P_n$, so that the current model can fit their target velocities. This connection can be formalized under regularity conditions, where the deviation in the regression solution induced by the auxiliary proposal is bounded by $(\alpha/\beta^2)\,D_{\mathrm{KL}}(Q_n\|P_n)$ (see App.~\ref{app:proposal-contribution}). 

An offline reference bank satisfies neither requirement: (i) auxiliary endpoints should receive rewards comparable to those of $P_n$, and (ii) they should remain close enough to $P_n$ for the current model to fit them. For (i), the rewards of bank endpoints stay fixed while current-model rewards improve, so the advantage increasingly reflects the source of an endpoint rather than its quality. For (ii), bank endpoints stay fixed while $P_n$ evolves, so they become increasingly difficult for the current model to fit. 
Section~\ref{sec:proposal-comparison} measures both quantities: the mean advantage and the velocity residual, and it shows that an offline reference bank lowers task reward. This result motivates an auxiliary proposal that follows $P_n$ while still introducing alternatives, which we construct next. App.~\ref{app:mixed-update} derives Eq.~\eqref{eq:main-nft-proposal-update} and states the corresponding conditions for AWM and RAM.

\subsection{Hybrid reference proposal: RefMix}
\label{sec:hybrid-proposal}

Figure~\ref{fig:refmix-pipeline} depicts the hybrid proposal used in RefMix. RefMix constructs auxiliary endpoints by applying the reference model during the early high-noise steps and the current model thereafter. Consider a sampler with $K$ solver steps ordered from high to low noise. Let $K_{\mathrm{ref}}\in\{0,\ldots,K\}$ be the number of reference steps and $k=K_{\mathrm{ref}}/K$ their fraction. With $t_j$ denoting the noise level at step $j$, the hybrid field is
\begin{equation}
\label{eq:hybrid-field-general}
v_{H_n^{(k)}}(x,t_j)=
\begin{cases}
v_{\mathrm{ref}}(x,t_j), & j\leq K_{\mathrm{ref}},\\
v_{\mathrm{old},n}(x,t_j), & j>K_{\mathrm{ref}}.
\end{cases}
\end{equation}
We write $H_n^{(k)}$ for the endpoint distribution produced by this sampler and set $Q_n=H_n^{(k)}$ in Eq.~\eqref{eq:proposal-mixture}. When $k=0$, the hybrid sampler is identical to $P_n$.

We restrict the reference model to the early steps because high-noise generation influences coarse layout and low-frequency structure \citep{balaji2022ediffi,sorokin2025imagerefl}. The reference prefix can place the trajectory in a different region, while the current-model suffix completes it using behavior acquired through reward tuning. 
Since hybrid and ordinary rollouts share the current-model suffix, $D_{\mathrm{KL}}(H_n^{(k)}\|P_n)$ cannot exceed the divergence between their states at the switch. 
Each RefMix hybrid rollout uses $K$ denoising evaluations, matching the cost of a standard rollout. RefMix therefore preserves the baseline rollout count and endpoint pool size while modifying only the endpoint proposal. Algorithm~\ref{alg:refmix-training} summarizes the overall update, with implementation details of the hybrid rollout provided in App.~\ref{app:experimental-details}. 

\vspace{-0.25cm}
\paragraph{Inference.}
The reference prefix can also be applied at inference, where increasing the reference fraction $k$ generally shifts the generation toward higher diversity at lower reward~\citep{sorokin2025imagerefl}. For RefMix, we use the same prefix fraction as during training, matching the hybrid trajectories used to construct the training endpoints. Importantly, the same inference intervention can be applied to the forward matching baselines, enabling a controlled comparison between baseline- and RefMix-trained models across $k$. To identify the most favorable reward--diversity trade-off, our analysis study in Section~\ref{sec:training-inference} compares the two trained models across $k$, disentangling the benefit of training-time endpoint proposal from reference-assisted inference. 


\begin{algorithm}[t]
\small
\caption{Reference-Assisted Endpoint Mixing}
\label{alg:refmix-training}
\begin{algorithmic}[1]
\Require Prompt $\Vec{c}$, pool size $N$, fraction $\alpha$, proposals $P_n,H_n$, reward model $\texttt{Reward}$, advantage rule $\mathcal A$
\State $N_Q\gets\alpha N$, \quad $N_P\gets N-N_Q$
\State Draw $N_P$ endpoints from $P_n(\cdot\mid \Vec{c})$
\State Draw $N_Q$ endpoints from $H_n(\cdot\mid \Vec{c})$
\State Combine the endpoints into $\{\Vec{x}_0^i\}_{i=1}^N$
\For{$i=1,\ldots,N$}
  \State $r_i\gets \texttt{Reward}(\Vec{x}_0^i,\Vec{c})$
\EndFor
\For{$i=1,\ldots,N$}
  \State $\Vec{A}_i\gets\mathcal A(r_i,\{r_j\}_{j=1}^N)$
  \State Draw $t_i$ and $\Vec{\epsilon}_i\sim\mathcal N(0,I)$
  \State Construct $(\Vec{x}_{t_i}^i,\Vec{u}_{t_i}^i)$ using Eq.~\eqref{eq:generic-analytic-noising}
\EndFor
\State Update $\Vec \theta$ using $\frac1N\sum_i\mathcal{L}_{\mathrm{FM}}^i(\Vec{\theta})$
\end{algorithmic}
\end{algorithm}

\section{Experiments}
\label{sec:experiments}

\begin{table*}[!t]
\centering
\small
\setlength{\tabcolsep}{4.0pt}
\renewcommand{\arraystretch}{0.9}
\newcommand{\cc}[1]{\cellcolor{cyan!10}#1}
\caption{
Quantitative evaluation across PickScore, GenEval, and OCR. ``+ RefMix'' uses $\alpha=0.75$.}
\vspace{-0.1cm}
\label{tab:main-results}
\resizebox{0.95\textwidth}{!}{
\begin{tabular}{llcccccccc}
\toprule
\multirow{2}{*}{\textbf{Benchmark}}
& \multirow{2}{*}{\textbf{Method}} & \multicolumn{1}{c}{\textbf{Target Reward}} & \multicolumn{4}{c}{\textbf{Visual
diversity}} & \multicolumn{3}{c}{\textbf{Auxiliary
rewards}} \\
\cmidrule(lr){3-3} \cmidrule(lr){4-7} \cmidrule(lr){8-10}
&
& \textbf{Reward}$\uparrow$ & $\DP\uparrow$ & $\DQ\uparrow$ & $\DLPIPS\uparrow$ & \textbf{Vendi}$\uparrow$
& \textbf{CLIPScore}$\uparrow$ &
\textbf{HPSv2}$\uparrow$ & \textbf{Aesthetic}$\uparrow$ \\
\midrule
\multirow{6}{*}{\textbf{PickScore}} & NFT & \textbf{22.56} & 0.084 & 0.078 & 0.390 & 1.596 & 0.264
& \textbf{0.316} & 6.490 \\
& \cc{NFT + RefMix} & \cc{22.54} & \cc{\textbf{0.099}} & \cc{\textbf{0.088}} & \cc{\textbf{0.429}} & \cc{\textbf{1.704}} & \cc{\textbf{0.266}}
& \cc{\textbf{0.316}} & \cc{\textbf{6.511}} \\
\cmidrule(lr){2-10}
& AWM & 22.56 & 0.083 & \textbf{0.079} & 0.566 & 1.589 & 0.262
& \textbf{0.319} & \textbf{6.633} \\
& \cc{AWM + RefMix} & \cc{\textbf{22.61}} & \cc{\textbf{0.087}} & \cc{0.077} & \cc{\textbf{0.570}} & \cc{\textbf{1.618}} & \cc{\textbf{0.264}}
& \cc{\textbf{0.319}} & \cc{6.600} \\
\cmidrule(lr){2-10}
& RAM & \textbf{22.77} & 0.077 & 0.074 & 0.592 & 1.549
& \textbf{0.283} &
\textbf{0.330} & 6.288 \\
& \cc{RAM + RefMix} & \cc{\textbf{22.77}} & \cc{\textbf{0.086}} & \cc{\textbf{0.081}} & \cc{\textbf{0.603}} & \cc{\textbf{1.613}} & \cc{0.281}
& \cc{0.326} & \cc{\textbf{6.315}} \\
\midrule
\multirow{6}{*}{\cellcolor{white}\textbf{GenEval}} & NFT & 0.781 & 0.084 & 0.081 & 0.381 & 1.574 & 0.286
& 0.261 & 5.201 \\
& \cc{NFT + RefMix} & \cc{\textbf{0.811}} & \cc{\textbf{0.087}} & \cc{\textbf{0.085}} & \cc{\textbf{0.460}} & \cc{\textbf{1.599}} & \cc{\textbf{0.291}}
& \cc{\textbf{0.262}} & \cc{\textbf{5.224}} \\
\cmidrule(lr){2-10}
& AWM & \textbf{0.815} & 0.049 & 0.048 & 0.340 & 1.338 & 0.289
& 0.268 & 5.147 \\
& \cc{AWM + RefMix} & \cc{0.799} & \cc{\textbf{0.132}} & \cc{\textbf{0.129}} & \cc{\textbf{0.618}} & \cc{\textbf{1.893}} &
\cc{\textbf{0.294}} &
\cc{\textbf{0.274}} & \cc{\textbf{5.442}} \\
\cmidrule(lr){2-10}
& RAM & 0.822 & 0.059 & 0.058 & 0.503 & 1.397 & 0.294
& 0.285 & 5.183 \\
& \cc{RAM + RefMix} & \cc{\textbf{0.837}} & \cc{\textbf{0.101}} & \cc{\textbf{0.099}} & \cc{\textbf{0.596}} & \cc{\textbf{1.690}} & \cc{\textbf{0.296}}
& \cc{\textbf{0.295}} & \cc{\textbf{5.377}} \\
\midrule
\multirow{6}{*}{\cellcolor{white}\textbf{OCR}} & NFT & 0.933 & 0.102 & 0.099 & \textbf{0.271} & 1.725 & \textbf{0.250}
& \textbf{0.135} & \textbf{3.768} \\
& \cc{NFT + RefMix} & \cc{\textbf{0.934}} & \cc{\textbf{0.114}} & \cc{\textbf{0.111}} & \cc{0.267} & \cc{\textbf{1.808}} & \cc{0.242}
& \cc{0.124} & \cc{3.575} \\
\cmidrule(lr){2-10}
& AWM & 0.932 & 0.157 & 0.157 & \textbf{0.548} & 2.192 & \textbf{0.279}
& \textbf{0.176} & \textbf{4.814} \\
& \cc{AWM + RefMix} & \cc{\textbf{0.933}} & \cc{\textbf{0.167}} & \cc{\textbf{0.168}} & \cc{0.484} & \cc{\textbf{2.276}} & \cc{0.271}
& \cc{0.148} & \cc{4.294} \\
\cmidrule(lr){2-10}
& RAM & 0.833 & 0.089 & 0.090 & 0.622 & 1.634 & \textbf{0.312}
& \textbf{0.272} & \textbf{5.100} \\
& \cc{RAM + RefMix} & \cc{\textbf{0.890}} & \cc{\textbf{0.115}} & \cc{\textbf{0.116}} & \cc{\textbf{0.640}} & \cc{\textbf{1.827}}
& \cc{0.303} &
\cc{0.250} & \cc{4.991} \\
\bottomrule
\end{tabular}
}
\vspace{-0.2cm}
\end{table*}

\begin{figure*}[t]
\centering
\includegraphics[width=0.95\textwidth]{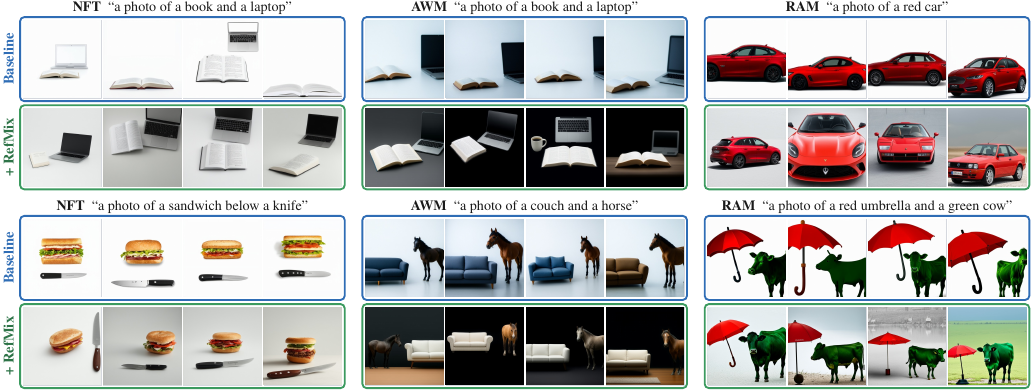}
\caption{Qualitative comparison on GenEval. Baseline samples converge to similar layouts and backgrounds, whereas RefMix retains alternatives in viewpoint and scene. All samples share initial noise.
}
\label{fig:qualitative-geneval}
\vspace{-0.3cm}
\end{figure*}

\begin{figure*}[t]
\centering
\vspace{-0.3cm}
\begin{minipage}[t]{0.49\textwidth}
\centering
\includegraphics[width=\linewidth]{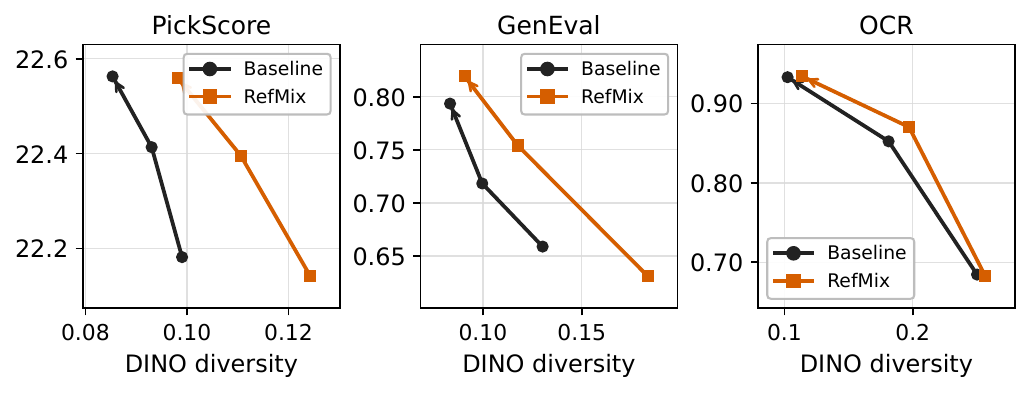}
\vspace{-0.7cm}
\captionof{figure}{Reward--diversity trajectories over NFT training.}
\label{fig:training-progression}
\end{minipage}
\hfill
\begin{minipage}[t]{0.49\textwidth}
\centering
\includegraphics[width=\linewidth]{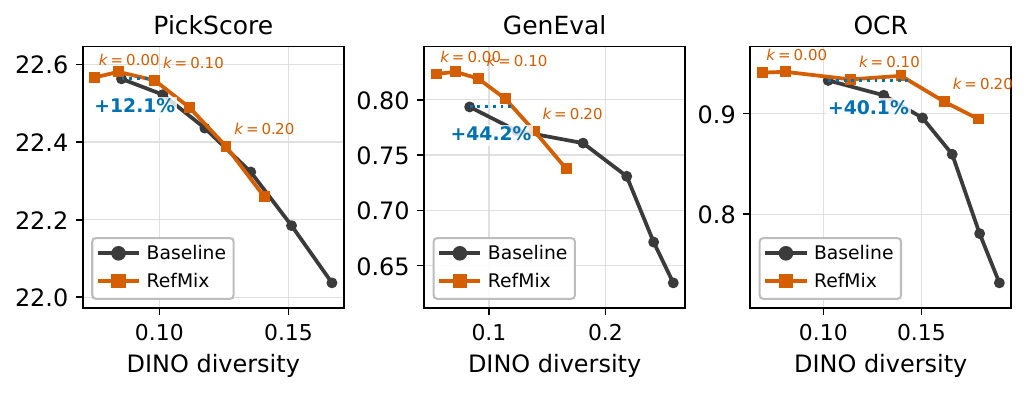}
\vspace{-0.7cm}
\captionof{figure}{NFT inference intervention across reference fractions $k$.}
\label{fig:inference-intervention}
\end{minipage}
\vspace{-0.4cm}
\end{figure*}

\subsection{Setup and evaluation protocol}

We fine-tune SD3.5-Medium \citep{esser2024scaling} with LoRA \citep{hu2021lora} on three rewards: PickScore \citep{kirstain2023pick}, GenEval \citep{ghosh2023geneval}, and OCR; the OCR reward follows FlowGRPO \citep{liu2025flowgrpo}. For each reward, we run three forward matching objectives, NFT \citep{zheng2025diffusionnft}, AWM \citep{xue2025awm}, and RAM \citep{bergmeister2026ram}, with and without RefMix. Alongside the optimized reward, we report CLIPScore \citep{hessel2021clipscore}, HPSv2 \citep{wu2023human}, and the Aesthetic score \citep{schuhmann2022laionaesthetics} as auxiliary evaluations of image quality and text alignment.

The existing baseline trains on current-model endpoints and uses the tuned field at every inference step. RefMix trains on the mixed pool and uses hybrid inference, with $\alpha=0.75$ and a nominal reference prefix of $10\%$; the realized fractions per objective are given in App.~\ref{app:experimental-details}. All comparisons use the same evaluation prompts and initial noise. Sections~\ref{sec:training-inference} and~\ref{sec:proposal-comparison} complement these full-pipeline results with controlled comparisons.

\paragraph{Diversity metrics.}

A higher reward can be reached by producing near-identical images for a prompt, so we report the optimized reward together with within-prompt visual diversity.
For $C$ evaluation samples per prompt, let $d_{ij}$ be the DINOv3 \citep{simeoni2025dinov3} pairwise cosine distance between samples $i$ and $j$. Our primary measure is the average distance between two samples drawn uniformly within a prompt,
\begin{equation}
\DP=\frac{1}{C(C-1)}\sum_{i\neq j}d_{ij}.
\label{eq:diversity-uniform}
\end{equation}
$\DLPIPS$ replaces $d_{ij}$ with the LPIPS \citep{zhang2018unreasonable} distance under the same convention. We also report the Vendi score \citep{friedman2022vendi} as an effective number of distinct samples within a prompt.

Since $\DP$ does not incorporate the rewards, it does not distinguish variation among the samples the reward ranks highest from variation among those it ranks lowest. We therefore define a reward-aware diversity $\DQ$, in which higher reward samples receive larger weight. For reward $r_i$,
\begin{equation}
  q_i(\tau)=\frac{\exp(r_i/\tau)}{\sum_{j}\exp(r_j/\tau)},
  \qquad
  \DQ=\frac{\sum_{i\neq j}q_i q_j\,d_{ij}}{\sum_{i\neq j}q_i q_j}.
  \label{eq:diversity-reward-aware}
\end{equation}
The temperature $\tau>0$ sets how sharply the weights concentrate on high-reward samples. Since a common temperature need not induce comparable concentration across prompts, we choose $\tau$ per prompt so that the effective sample size $\ESS(q)=1/\sum_i q_i^2$ meets the target $\ESS(q)/C=0.25$. When tied maximum rewards make this target unattainable, the weights become uniform over the tied samples, so the achieved ESS can exceed the target.

\subsection{Main results}
\label{sec:main-results}

Table~\ref{tab:main-results} compares the full pipelines at the final evaluated checkpoint of each run. On PickScore, target rewards remain within $0.3\%$ of the corresponding baselines. DINO diversity $\DP$, LPIPS diversity, and the Vendi score increase for all three objectives. Reward-aware diversity $\DQ$ increases for NFT and RAM, while AWM shows a small decrease.

On GenEval, the effect depends more strongly on the forward objective. NFT improves target reward from $0.781$ to $0.811$, and RAM from $0.822$ to $0.837$; both also improve all four diversity measures. AWM instead exhibits a reward--diversity trade-off: its reward decreases from $0.815$ to $0.799$, while visual diversity measures increase substantially. 
 
On OCR, NFT and AWM retain nearly unchanged target rewards, whereas RAM improves from $0.833$ to $0.890$. All three forward matching objectives increase $\DP$ and $\DQ$. However, LPIPS decreases for NFT and AWM, and their auxiliary rewards also decline. RAM improves all four diversity measures, but its auxiliary rewards are lower than the baseline. These differences show that higher DINO diversity does not imply improvement in every perceptual measure or evaluator.

\paragraph{Qualitative comparison.}
Figure~\ref{fig:qualitative-geneval} shows qualitative examples on GenEval. Baseline samples share similar layouts and backgrounds, while RefMix produces changes in viewpoint, object appearance, and scene. In the AWM examples, much of the variation is in lighting and background. In the NFT book--laptop example, two of the four baseline samples omit the laptop, whereas every RefMix sample contains both objects from a different viewpoint. 

\paragraph{Training Dynamics.}
Figure~\ref{fig:training-progression} follows the on-policy baseline (NFT) and the RefMix pipeline through the evaluated checkpoints. Across all rewards, reward increases as DINO diversity decreases during training. RefMix trajectory stays at higher diversity for comparable reward, thus the gap in Table~\ref{tab:main-results} is not specific to the final checkpoint.

\begin{figure}[t]
\centering
\includegraphics[width=0.65\linewidth]{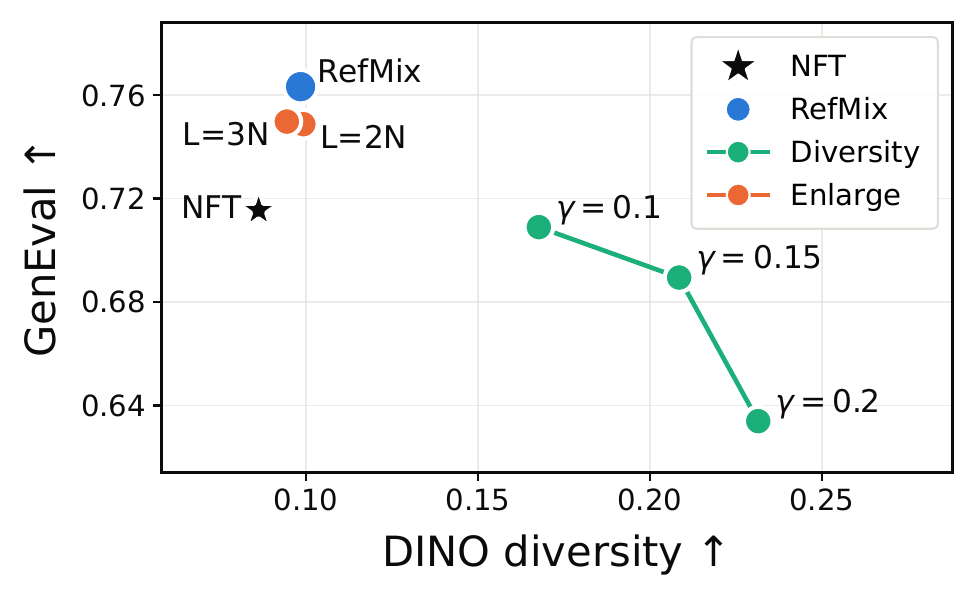}
\vspace{-0.3cm}
\caption{Reward--diversity comparison of NFT interventions on GenEval. Diversity adds a diversity advantage with weight $\gamma$; Enlarge selects $N$ endpoints from an on-policy pool of size $L$.}
\label{fig:diagnostic-interventions}
\end{figure}

\begin{figure*}[t]
\centering
\includegraphics[width=0.95\textwidth]{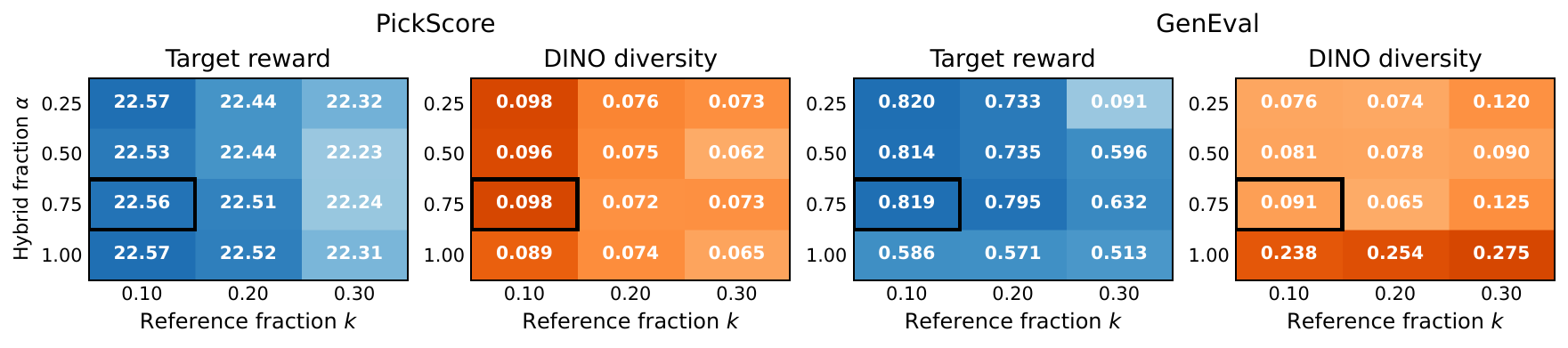}
\vspace{-0.4cm}
\caption{Joint ablation of the hybrid fraction $\alpha$ and the training reference fraction $k$. Black boxes mark the main setting
$(\alpha,k)=(0.75,0.10)$. All cells use NFT at update 30 and $k=0.10$ at inference.}
\vspace{-0.4cm}
\label{fig:proposal-ablation}
\end{figure*}

\subsection{Separating training and inference effects}
\label{sec:training-inference}

RefMix alters the endpoint proposal during training and uses the corresponding hybrid sampler at inference. To separate these effects, we hold each trained NFT
checkpoint fixed and vary only the fraction $k$ of inference calls assigned to the reference prefix. Applying this intervention to the baseline checkpoint tests inference-only composition, following \citep{sorokin2025imagerefl}.

Figure~\ref{fig:inference-intervention} shows that increasing $k$ generally moves either fixed checkpoint toward greater diversity at lower reward. Changing the training proposal can improve this trade-off: at the reward of baseline inference, the interpolated RefMix curves provide $12.1\%$, $44.2\%$, and $40.1\%$ more DINO diversity on PickScore, GenEval, and OCR, respectively. The PickScore curves intersect, whereas GenEval and OCR show a clearer advantage in the high-reward range. Under the shared $k=0.10$ rule, RefMix also attains higher reward on all three benchmarks. AWM and RAM show the same pattern as shown in Figure~\ref{fig:inference-awm-ram}. Thus, the observed benefit is not explained solely by inserting reference calls at inference.

\subsection{Alternative endpoint proposals $Q_n$}
\label{sec:proposal-comparison}

\paragraph{Interventions within the on-policy pool.}
\label{sec:on-policy-diagnostics}
We first compare two interventions that retain current-model generation: adding an explicit diversity advantage with weight $\gamma$~\citep{he2025gardo} and selecting $N$ endpoints from an enlarged on-policy pool, inspired by \citep{liu2026drift}. In the NFT GenEval comparison in Figure~\ref{fig:diagnostic-interventions}, diversity shaping increases variation but reduces task reward, whereas enlarged-pool selection improves both at additional generation cost. These observations motivate changing the endpoint source without establishing a general limitation of diversity regularization. Please see App.~\ref{app:diagnostic-interventions} for the settings and selection rule.

\paragraph{Offline reference bank.}
The formulation in Eq.~\eqref{eq:proposal-mixture} enables a direct comparison between an offline reference bank and RefMix while keeping the task reward and NFT update unmodified. This comparison is designed to test whether reusing reference endpoints is sufficient, or whether completing auxiliary endpoints with the current model changes the reward--diversity trade-off. For each training prompt, the bank holds $B=48$ reference-model endpoints; each group draws its $\alpha N =18$ auxiliary endpoints from the bank without replacement, and a top-50\% variant draws only from the $24$ endpoints with the highest reward at construction. Bank endpoints are rescored at every update, and their advantages are computed jointly with the current-model endpoints. All proposals use the same pool size, update schedule, and auxiliary fraction, and all models are evaluated using current-model-only inference. Both bank variants reduce task reward on all three tasks as shown in Table~\ref{tab:proposal-comparison}, and only the top-50\% bank on GenEval raises diversity above the baseline. Reusing reference endpoints therefore does not provide the diversity that motivates the bank. The online RefMix proposal instead retains or improves the baseline reward on all three tasks. 

\begin{figure}[t]
\vspace{-0.35cm}
\centering
\includegraphics[width=0.9\linewidth]{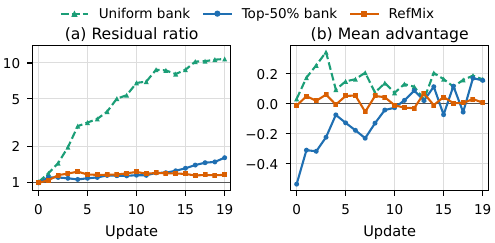}
\vspace{-0.3cm}
\caption{Source-wise diagnostics with NFT on GenEval. (a) RMS velocity residual on auxiliary endpoints relative to current-model endpoints. (b) Mean advantage of current-model endpoints.}
\vspace{-0.4cm}
\label{fig:source-diagnostics}
\end{figure}

Further, we analyze both bank variants and the performance gap. Figure~\ref{fig:source-diagnostics} (a) compares how well the current model fits auxiliary and current-model endpoints. The velocity residual for the uniform bank grows to nearly 10$\times$ that of current-model endpoints, while the hybrid proposal remains within 16\% throughout training. Next, we examine the mean advantage of current-model endpoints, where values near zero indicate balanced weighting across proposal sources as shown in Figure~\ref{fig:source-diagnostics} (b). RefMix remains close to zero, whereas the uniform bank consistently favors current-model endpoints and the top-50\% bank initially favors bank endpoints. Overall, the fixed banks fail to maintain both model compatibility and balanced reward weighting, while RefMix preserves both properties throughout training.

\begin{table}[t]
\centering
\large
\setlength{\tabcolsep}{4pt}
\caption{Effect of the training endpoint proposal on reward and diversity with NFT, under current-model-only inference. Parentheses give the change relative to the baseline.}
\vspace{-0.2cm}
\label{tab:proposal-comparison}
\setlength{\tabcolsep}{3pt}
\newcommand{\dn}[1]{\hspace{0.2pt}\raisebox{-1.6pt}{\small\color{BrickRed}(-#1)}}
\newcommand{\up}[1]{\hspace{0.2pt}\raisebox{-1.6pt}{\small\color{ForestGreen}(+#1)}}
\resizebox{\linewidth}{!}{%
\begin{tabular}{lcccccc}
\toprule
\multirow{2}{*}{\textbf{Proposal}} & \multicolumn{2}{c}{\textbf{PickScore}} & \multicolumn{2}{c}{\textbf{GenEval}} & \multicolumn{2}{c}{\textbf{OCR}} \\
\cmidrule(lr){2-3}\cmidrule(lr){4-5}\cmidrule(lr){6-7}
 & \textbf{Rew.}$\uparrow$ & $\DP\uparrow$ & \textbf{Rew.}$\uparrow$ & $\DP\uparrow$ & \textbf{Rew.}$\uparrow$ & $\DP\uparrow$ \\
\midrule
Baseline & 22.41 & 0.093 & 0.718 & 0.100 & 0.933 & 0.102 \\
Bank (uniform) & 19.70\dn{2.71} & 0.079\dn{.014} & 0.334\dn{.384} & 0.096\dn{.004} & 0.000\dn{.933} & 0.023\dn{.079} \\
Bank (top-50\%) & 20.35\dn{2.06} & 0.080\dn{.013} & 0.583\dn{.135} & 0.181\up{.081} & 0.100\dn{.833} & 0.083\dn{.019} \\
RefMix & 22.42\up{0.01} & 0.085\dn{.008} & 0.769\up{.051} & 0.076\dn{.024} & 0.942\up{.009} & 0.069\dn{.033} \\
\bottomrule
\end{tabular}}
\vspace{-0.2cm}
\end{table}

\subsection{Ablations}
\label{sec:ablations}

\paragraph{Training endpoint proposal composition.}
We jointly vary the hybrid endpoint fraction $\alpha$ and reference-prefix fraction $k$ with NFT to study the training proposal composition. All cells in Figure~\ref{fig:proposal-ablation} use update 30 and the same $k=0.10$ hybrid inference rule. The matrix consequently
isolates the composition and reference-prefix length of the training proposal. It compares $\alpha\in\{0.25,0.50,0.75,1.00\}$ and $k\in\{0.10,0.20,0.30\}$. On PickScore, $k=0.10$ gives higher reward and DINO diversity than the longer prefixes. GenEval exhibits a stronger interaction: at $k=0.10$, the mixed proposals with $\alpha<1$ achieve rewards of $0.814$--$0.820$, whereas replacing the entire proposal ($\alpha=1$) increases DINO diversity to $0.238$ but reduces reward to $0.586$. Longer prefixes do not consistently improve the trade-off; for example, $(\alpha,k)=(0.75,0.20)$ lowers both reward and diversity relative to $(0.75,0.10)$. These results support our main setting of a  short reference prefix while retaining current model endpoints, $(\alpha,k)=(0.75,0.10)$.

\section{Conclusion}

We studied reference-assisted endpoint proposals for online forward matching, decoupling training endpoint generation from the reward update. Current-model endpoints restrict mode exposure, while auxiliary proposals can recover underrepresented modes. Effective auxiliary endpoints must remain compatible with the evolving model and maintain comparable rewards, conditions that fixed reference banks fail to satisfy. RefMix addresses these limitations through a reference prefix followed by a current-model suffix, while keeping the original forward matching update. Across NFT, AWM, and RAM, RefMix consistently improves diversity while preserving task rewards. At matched reward, RefMix remains more diverse than baselines under the same reference-assisted inference, highlighting the benefit of diversifying endpoints during training.

\subsection*{LLM Usage Statement}

We used Codex and Claude to edit manuscript text and check consistency between the text, figures, and tables; to review and debug training and evaluation code; to write the plotting scripts that generate several figures from our logged results; and to search for related work and give feedback on the method and experimental design. All model outputs and numerical results were produced by our experimental pipelines, not generated by AI tools. The authors take full responsibility for the content.

{
    \small
    \bibliographystyle{ieeenat_fullname}
    \bibliography{main}
}

\clearpage
\maketitlesupplementary
\appendix

\section{Experimental Details}
\label{app:experimental-details}

\paragraph{Common training settings.}
We fine-tune with rank-32 LoRA adapters and LoRA scaling parameter 64. All main runs use a learning rate of $3\times10^{-4}$ and weight decay $10^{-4}$. The NFT branch coefficient $\beta$ is $1.0$ for PickScore and GenEval and $0.1$ for OCR, following the released configuration. The training CFG scale is $1.0$ for NFT and AWM and $2.0$ for RAM. Each update uses $N=24$ endpoints per prompt. NFT and RAM use 48 prompts per update, whereas AWM follows its released configuration: 24 prompts for PickScore and 72 for GenEval and OCR. NFT rollouts use 10 deterministic DPM2 steps; AWM rollouts use 14 steps of a second-order SA-Solver predictor--corrector without added noise, and RAM rollouts use 20 deterministic Euler steps. Table~\ref{tab:prefix-protocol} collects the per-method and per-task settings.

\paragraph{Hybrid proposal protocol.}
For each prompt group of $N=24$ endpoints, the baseline draws all endpoints from the current model. With $\alpha=0.75$, RefMix draws $N_Q=\alpha N=18$ hybrid endpoints and $N_P=6$ current-model endpoints, scores all 24 with the task reward, and computes their advantages within the group. The reference prefix covers the first call of ten for NFT, the first of fourteen for AWM, and the first two of twenty for RAM (Table~\ref{tab:prefix-protocol}). We write $k_{\mathrm{train}}$ and $k_{\mathrm{test}}$ for the reference fraction $k$ during training and at inference. The realized training fraction is thus $k_{\mathrm{train}}=0.1$ for NFT and RAM and $k_{\mathrm{train}}=1/14$ for AWM. Since the reference model is the backbone with the LoRA adapter disabled, each reference call replaces a current-model call on the same backbone. A hybrid rollout therefore uses the same $K$ calls as an ordinary rollout and requires no second copy of the model.

\begin{table}[t]
\centering
\small
\setlength{\tabcolsep}{3pt}
\caption{Per-method and per-task settings. Coef.\ is the method-specific coefficient: the branch coefficient $\beta$ for NFT, $\beta_{\mathrm{ref}}$ for AWM, and reward coefficient for RAM. Updates is the update count of the checkpoint reported in Table~\ref{tab:main-results}.}
\label{tab:prefix-protocol}
\resizebox{\linewidth}{!}{%
\begin{tabular}{llcccccc}
\toprule
Method & Task & \text{Train } $K_{\mathrm{ref}}/K$ & $k_{\mathrm{train}}$ & $\text{Test } K_{\mathrm{ref}}/K$ & Coef. & train/test CFG & Updates \\
\midrule
\multirow{3}{*}{NFT} & PickScore & 1/10 & 0.10 & 4/40 & 1.0 & 1.0/1.0 & 30 \\
& GenEval & 1/10 & 0.10 & 4/40 & 1.0 & 1.0/1.0 & 30 \\
& OCR & 1/10 & 0.10 & 4/40 & 0.1 & 1.0/1.0 & 20 \\
\midrule
\multirow{3}{*}{AWM} & PickScore & 1/14 & 0.071 & 3/40 & $10^{-4}$ & 1.0/1.0 & 28 \\
& GenEval & 1/14 & 0.071 & 3/40 & $10^{-3}$ & 1.0/1.0 & 28 \\
& OCR & 1/14 & 0.071 & 3/40 & $3\times10^{-3}$ & 1.0/1.0 & 18 \\
\midrule
\multirow{3}{*}{RAM} & PickScore & 2/20 & 0.10 & 4/40 & 1000 & 2.0/2.0 & 30 \\
& GenEval & 2/20 & 0.10 & 4/40 & 100 & 2.0/4.5 & 30 \\
& OCR & 2/20 & 0.10 & 4/40 & 100 & 2.0/4.5 & 20 \\
\bottomrule
\end{tabular}}
\end{table}

\paragraph{Offline-bank protocol.}
\label{app:offline-bank}
For the NFT comparisons in Table~\ref{tab:proposal-comparison}, we precompute $B=48$ reference-model endpoints for each training prompt with 10 deterministic DPM2 steps and CFG scale $1.0$, and store the clean latent and decoded image of each endpoint. At each update, the uniform variant draws the $\alpha N=18$ auxiliary endpoints without replacement from all 48 bank endpoints, and the top-50\% variant draws them from the 24 bank endpoints with the highest reward at construction. These are combined with $N_P=6$ fresh current-model endpoints. Bank endpoints are rescored at every update, and advantages are computed jointly over the combined group.

\paragraph{Inference protocol.}
Target reward is evaluated with one generation for every held-out prompt: 2,048 prompts for PickScore, 2,211 for GenEval, and 1,017 for OCR. Evaluation uses 40 solver calls and images at $512\times512$ resolution. The CFG scale is $1.0$ for NFT and AWM; RAM uses $2.0$ on PickScore and $4.5$ on GenEval and OCR. RefMix uses hybrid inference, with the reference model for the first calls and the trained model for the rest: four of 40 calls for NFT and RAM, and three for AWM, the closest count to its training fraction of $1/14$. Baselines use the trained model at every call.

\begin{figure*}[t]
\centering
\includegraphics[width=\textwidth]{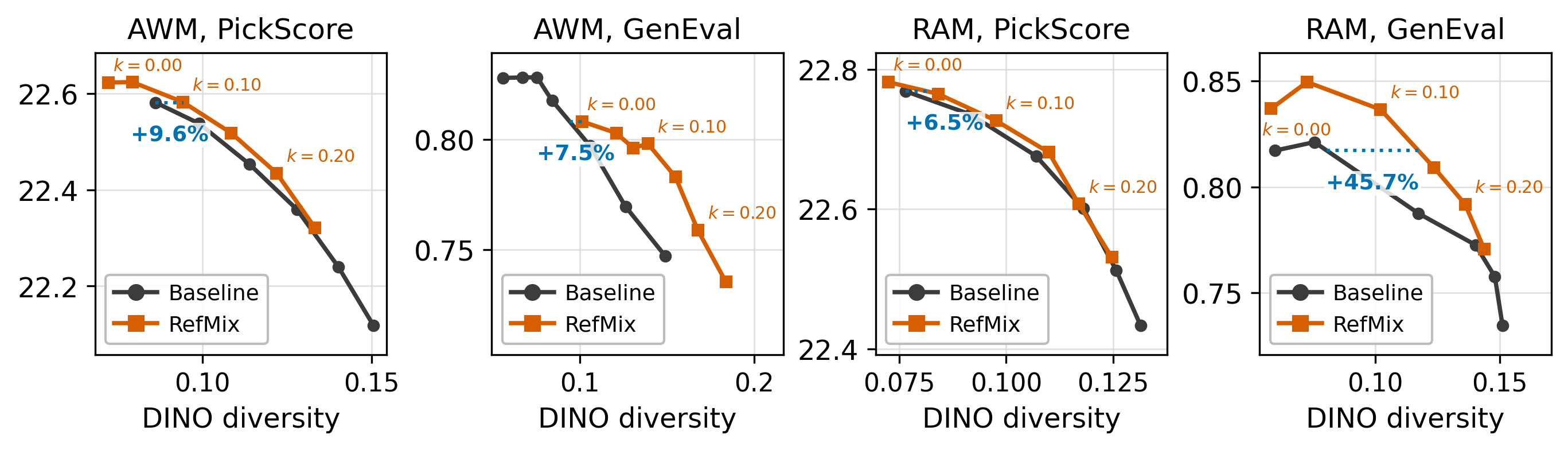}
\vspace{-0.6cm}
\caption{Inference intervention at fixed AWM and RAM checkpoints on PickScore and GenEval, varying the inference reference fraction $k_{\mathrm{test}}$. Percentages give the $\DP$ gain of RefMix at the matched reward (App.~\ref{app:inference-control}).}
\label{fig:inference-awm-ram}
\end{figure*}

\paragraph{Within-prompt diversity.}
The main quantitative diversity evaluation uses
a fixed panel of 64 held-out prompts with $C=16$ generations per prompt. $\DP$ and $\DLPIPS$ average over distinct sample pairs, as in Eq.~\eqref{eq:diversity-uniform}, and all quantities are computed within each prompt before averaging across prompts.

\paragraph{Controlled inference sweeps.}
\label{app:inference-control}
In Figure~\ref{fig:inference-intervention}, both baseline-trained and RefMix-trained checkpoints use $K_{\mathrm{ref}}\in\{0,2,4,6,8,10\}$ out of 40 inference calls, i.e., $k_{\mathrm{test}}\in\{0,0.05,0.10,0.15,0.20,0.25\}$. Prompt sets, initial-noise seeds, CFG scale, and the evaluation implementation are shared within each task's sweep. We obtain the $\DP$ of RefMix at this reward by linear interpolation between the two adjacent evaluated points on the RefMix curve.

We repeat the inference sweep for AWM and RAM on PickScore and GenEval in Figure~\ref{fig:inference-awm-ram}. For RAM on GenEval, RefMix attains higher reward at every evaluated $k_{\mathrm{test}}$ and provides $45.7\%$ more $\DP$ at the matched reward. On PickScore, the RefMix curve lies above the baseline curve, with $9.6\%$ more $\DP$ for AWM and $6.5\%$ for RAM at the matched reward. For AWM on GenEval, RefMix does not reach the highest baseline reward. Over the reward range where the two curves overlap, the RefMix curve still lies above the baseline curve, with $7.5\%$ more $\DP$.

\paragraph{Source diagnostics and bank interventions.}
\label{app:bank-diagnostics}

We measure the two source-wise quantities of Section~\ref{sec:proposal-comparison}: the velocity residual and the mean advantage. For each sampled endpoint and training timestep, the velocity residual is the per-sample RMS residual $e_i = \|\Vec{v}_\theta(\Vec{x}_t^i, t) - \Vec{u}_t^i\|_2 / \sqrt{d_{\mathrm{lat}}}$, where $d_{\mathrm{lat}}$ denotes the latent dimension, and we average $e_i$ within each source. Figure~\ref{fig:source-diagnostics}(a) reports the ratio of this average on auxiliary endpoints to that on current-model endpoints. Figure~\ref{fig:source-diagnostics}(b) reports the mean advantage of current-model endpoints at each update.

To test whether the larger velocity residual or the nonzero mean advantage in Figure~\ref{fig:source-diagnostics} explains the lower task reward of the offline reference bank, we apply two interventions to the top-50\% bank. First, the refresh control targets the velocity residual: starting from the same top-50\% bank, it regenerates $48$ candidates per prompt with the current model every five updates and keeps the $24$ with the highest reward for the next five updates. Second, the source-centering control targets the mean advantage: on top of the refresh control, it removes the mean advantage of each source after prompt-group standardization. Let $A_i$ denote the standardized advantage and $\mathcal{G}_{\Vec{c},s}$ denote the endpoints for prompt $\Vec{c}$ from source $s$. Before the advantages enter the NFT loss, we apply
\begin{equation}
A_i \leftarrow A_i - \frac{1}{|\mathcal{G}_{\Vec{c},s}|}\sum_{j\in\mathcal{G}_{\Vec{c},s}} A_j,
\qquad i \in \mathcal{G}_{\Vec{c},s},
\end{equation}
which sets the mean advantage of each source to zero and leaves the variance within each source unchanged.

\begin{table}[t]
\centering
\small
\setlength{\tabcolsep}{4pt}
\caption{NFT bank interventions on GenEval. Res.\ ratio is the ratio of the RMS velocity residual on auxiliary endpoints to that on current-model endpoints at the final update. $|\bar A_{\mathrm{cur}}|$ is the absolute mean advantage of current-model endpoints.}
\label{tab:bank-interventions}
\begin{tabular}{lccc}
\toprule
Training configuration & Reward $\uparrow$ & Res.\ ratio & $|\bar A_{\mathrm{cur}}|$ \\
\midrule
Current-model baseline & 0.718 & -- & -- \\
Uniform bank & 0.334 & 10.80 & 0.15 \\
Top-50\% bank & 0.583 & 1.61 & 0.15 \\
\quad + refresh every 5 updates & 0.565 & 1.02 & 0.43 \\
\qquad + source centering & 0.684 & 1.04 & 0.00 \\
\bottomrule
\end{tabular}
\end{table}

Table~\ref{tab:bank-interventions} reports the results. Refreshing the reference bank brings the residual ratio to $1.02$, against $1.61$ for the fixed top-50\% bank, yet reward does not recover ($0.583$ to $0.565$). Meanwhile, $|\bar A_{\mathrm{cur}}|$ increases from $0.15$ to $0.43$, consistent with the refresh rule favoring high-reward bank endpoints. Setting the mean advantage of each source to zero with the source-centering control raises reward to $0.684$, still below the current-model baseline of $0.718$. Matching the velocity residual is therefore not sufficient while the mean advantage differs between sources, consistent with the two requirements in Section~\ref{sec:proposal-analysis}.

\section{On-Policy Amplification and Toy Example} 
\label{app:amplification-toy}

Forward matching retains a supervised noising-and-regression structure, and its distribution of training endpoints $P_{n}$ come from the current model. After each update, the model defines the next endpoint distribution $P_{n+1}(\cdot\mid \Vec{c})$ and generates the next endpoint pool, so training forms an on-policy feedback loop through its own samples. To illustrate the local effect of this feedback, we consider an idealized target analyzed by AdvantageFlow \citep{kveton2026advantageflow}:
\begin{equation}
P_{n+1}(\Vec{x}_0\mid
\Vec{c}) \propto \bigl(1+\eta A(\Vec{x}_0,\Vec{c})\bigr)P_n(\Vec{x}_0\mid \Vec{c}),
\end{equation}
where $A$ is an advantage function with $\mathbb{E}_{P_n}[A(\cdot,\Vec{c})]=0$, $\eta>0$ is a step size, and $1+\eta A(\Vec{x}_0,\Vec{c})\geq0$.
For endpoints $\Vec{x},\Vec{y}$ with positive density and $1+\eta A(\Vec{y},\Vec{c})>0$, this update changes the target density ratio according to
\begin{equation}
\frac{P_{n+1}(\Vec{x}\mid \Vec{c})}{P_{n+1}(\Vec{y}\mid \Vec{c})}
=\frac{1+\eta
A(\Vec{x},\Vec{c})}{1+\eta A(\Vec{y},\Vec{c})}
\cdot \frac{P_n(\Vec{x}\mid \Vec{c})}{P_n(\Vec{y}\mid \Vec{c})} .
\label{eq:relative-amplification}
\end{equation}
If the model trains to reproduce this target, the density ratio increases whenever $A(\Vec{x},\Vec{c})>A(\Vec{y},\Vec{c})$, so repeated on-policy forward matching can concentrate the endpoint distribution on high-advantage endpoints. Eq.~\eqref{eq:relative-amplification} leaves the ratio between two endpoints with equal advantage unchanged. With a finite pool, however, the less probable mode is observed less often (Section~\ref{sec:diagnosis}) and can receive less direct supervision.

\paragraph{Toy setup.}
Figure~\ref{fig:toy-contraction} shows this contraction in a controlled multimodal flow with known data modes and reward landscape. The data are nine equal-weight Gaussian modes on a line, and the reward is high only on the two end modes. The reference model is a small MLP flow sampled with 10 Euler steps, which NFT fine-tunes for 1,500 updates. RefMix uses $\alpha=0.75$ and the reference model for the first 4 of the 10 steps, i.e., $k=0.4$. The Vendi score in Figure~\ref{fig:toy-contraction} is computed from 4,000 samples with an RBF kernel of bandwidth 1. As a result, the reference model covers all nine modes, including both high-reward modes. NFT increases reward but sends most trajectories to one of the two high-reward modes, whereas RefMix keeps both. 

\section{Diagnostic Intervention Details}
\label{app:diagnostic-interventions}

This section details the two on-policy interventions of Section~\ref{sec:proposal-comparison}, explicit diversity shaping and enlarged-pool selection. All diagnostics use NFT on GenEval, with $N=24$ endpoints per prompt retained for optimization, and share the task reward, batch size, and evaluation protocols. Figure~\ref{fig:diagnostic-interventions} compares the resulting models. 

\paragraph{Explicit diversity shaping.}

Inspired by GARDO~\citep{he2025gardo}, we keep the baseline pool of $N$ endpoints from $P_n$ and add a diversity advantage. Each endpoint receives a diversity score that increases as it becomes less similar to the other endpoints of its prompt group, measured with an RBF kernel on $\ell_2$-normalized DINOv3 features. The task reward and this diversity score are standardized separately within each prompt group, producing $A_i$ and $A_i^{\mathrm{div}}$, and are combined as
\begin{equation}
\widetilde A_i
=
A_i+\gamma A_i^{\mathrm{div}}.
\end{equation}
We evaluate $\gamma\in\{0.1,0.15,0.2\}$.

\paragraph{Enlarged on-policy pool.}

Inspired by DRIFT~\citep{liu2026drift}, we draw $L\in\{2N,3N\}$ on-policy candidates and retain $N$ for optimization, which uses $L/N$ times the rollout budget at the same optimization batch size. We adopt the selection rule of DRIFT: the retained group is the $N$-point nearest-neighbor neighborhood in DINOv3 feature space with the smallest summed distance to its center. We select
\begin{equation}
i^\star
=
\argmin_i
\sum_{j\in\mathrm{nn}_N(i)}
\lVert\Vec{\phi}_i-\Vec{\phi}_j\rVert_2,
\qquad
\mathcal I^\star=\mathrm{nn}_N(i^\star),
\end{equation}
where $\Vec{\phi}$ denotes DINO feature. The retained group $\mathcal I^\star$ is the $N$-point nearest-neighbor neighborhood with the smallest summed distance to its center. 

\subsection{Candidate exposure and enlarged-pool cost}

Section~\ref{sec:online-sampling} states that a high-reward region with small probability under the current model rarely appears in a pool of $N$ endpoints. We derive the probability that such a region appears at least once in the pool, and then invert it to obtain the idealized pool size.

Fix a prompt $\Vec{c}$ and a high-reward region
$S$ that is underrepresented in the current model, and write $p=p_n(S\mid \Vec{c})$. Let $\Vec{x}_0^1,\ldots,\Vec{x}_0^N$ be independent draws from $P_n(\cdot\mid \Vec{c})$. A single draw lies outside $S$ with probability $1-p$. By independence,
\begin{align}
\Pr\!\left(\Vec{x}_0^i\notin S\text{ for all }i\le N\mid \Vec{c}\right)
&=\prod_{i=1}^{N}\Pr(\Vec{x}_0^i\notin S\mid \Vec{c})\\&=(1-p)^N.
\end{align}
Taking the complementary yields
\begin{equation}
\Pr\!\left(\exists i\le N:\Vec{x}_0^i\in S\mid \Vec{c}\right)
=1-(1-p)^N.
\label{eq:on-policy-exposure}
\end{equation}
We call this probability the exposure of $S$. For the enlarged pool, replacing $N$ by $L$ gives
\begin{equation}
\Pr\!\left(\exists i\leq L:\Vec{x}_0^i\in S\mid \Vec{c}\right)
=1-(1-p)^L.
\label{eq:candidate-exposure}
\end{equation}
For $0<\delta<1$ and $0<p<1$, requiring this probability to be at least $1-\delta$ is equivalent to
\begin{equation*}
(1-p)^L\leq\delta
\quad\Longrightarrow\quad
L\log(1-p)\leq\log\delta.
\end{equation*}
The logarithm multiplying $L$ is negative, so division reverses the inequality:
\begin{equation}
L\geq\frac{\log\delta}{\log(1-p)}
\approx\frac{\log(1/\delta)}{p}.
\label{eq:exposure-cost}
\end{equation}
The approximation uses $\log(1-p)\approx-p$ for small positive $p$. Therefore, the required endpoints for a fixed exposure scales as $1/p$, and the enlarged pool becomes more required as reward tuning lowers the probability $p$ that the current model assigns to $S$. RefMix instead raises exposure at a fixed pool size $N$ when $H_n^{(k)}$ places more mass on $S$ than $P_n$ intrinsically (App.~\ref{app:discovery-discrepancy}).

\begin{table*}[t]
\centering
\small
\caption{Coefficients, anchors, and the relation of each mapping onto
Eq.~(\ref{eq:App.-rvm-loss}).}
\label{tab:forward matching-views}
\begin{tabular}{lllll}
\toprule
Method
& $\lambda_i(t,A_i)$ & $g_i(t,A_i)$ & $\Vec{v}_{\texttt{anc}}$ & Relation \\
\midrule
NFT & $\beta A_i$ & $\beta(\beta-A_i)$ & $\Vec{v}_{\mathrm{old},n}$
& Loss, up to a $\theta$-independent term \\
AWM
& $\rho_i A_i w_{\mathrm{pol}}(t)$
& $\beta_{\mathrm{AWM}}w_{\mathrm{KL}}(t)$
& $\Vec{v}_{\mathrm{ref}}$ or $\Vec{v}_{\mathrm{EMA}}$
& Gradient, at a fixed ratio $\rho_i$
\\
RAM & $r_i^{\mathrm{RAM}}$ & $1$ & $\Vec{v}_{\mathrm{ref}}$
& Gradient, at a fixed detached target \\
\bottomrule
\vspace{-0.4cm}
\end{tabular}
\end{table*}

\section{A Common Regression Form for Forward Matching Updates}
\label{app:forward-objectives}

We use reward-based velocity matching (RVM) \citep{choi2026scaling} to describe the common anchored-regression structure of NFT, AWM, and RAM. The time and prompt arguments are omitted when shared. The derivations describe the common regression form in Section~\ref{sec:forward-reward-setup}.

\paragraph{RVM as a common regression form.}
Following the RVM, write the per-endpoint anchored regression from Eq.~(\ref{eq:anchored-forward-regression}) as
\begin{align}
\mathcal{L}_{\mathrm{FM}}^i
={}&\lambda_i(t,A_i)\|\Vec{v}_\theta(\Vec{x}_t^i,t)-\Vec{u}_t^i\|^2 \nonumber\\
&+g_i(t,A_i)\|\Vec{v}_\theta(\Vec{x}_t^i,t)-\Vec{v}_{\texttt{anc}}(\Vec{x}_t^i,t)\|^2.
\label{eq:App.-rvm-loss}
\end{align}
For readability, write $\lambda_i$ and $g_i$ for these coefficients when their arguments are shared. At a fixed sample, the gradient with respect to the predicted velocity $\Vec{v}$ is
\begin{equation}
\nabla_{\Vec{v}}\mathcal{L}_{\mathrm{FM}}^i
=2(\lambda_i+g_i)(\Vec{v}-\Vec{v}_{\texttt{anc}})
-2\lambda_i(\Vec{u}_t^i-\Vec{v}_{\texttt{anc}}).
\label{eq:rvm-common-gradient}
\end{equation}
The sum $\lambda_i+g_i$ controls the quadratic curvature, while $\lambda_i$ controls the reward-weighted endpoint residual. The three methods reach Eq.~(\ref{eq:App.-rvm-loss}) or Eq.~(\ref{eq:rvm-common-gradient}) at different strengths, and Table~\ref{tab:forward matching-views} summarizes them.

\paragraph{NFT.}
Assume $\beta>0$, where $\beta$ is the NFT branch coefficient. For sample $i$, let $\omega_i\in[0,1]$ be the NFT reward weight and set $A_i=2\omega_i-1$, following NFT \citep{zheng2025diffusionnft}. The two implicit branches are
\begin{align}
\Vec{v}_{\theta,i}^{+}
&=(1-\beta)\Vec{v}_{\mathrm{old},n}(\Vec{x}_t^i,t)+\beta \Vec{v}_\theta(\Vec{x}_t^i,t), \nonumber\\
\Vec{v}_{\theta,i}^{-}
&=(1+\beta)\Vec{v}_{\mathrm{old},n}(\Vec{x}_t^i,t)-\beta \Vec{v}_\theta(\Vec{x}_t^i,t).
\end{align}
The per-endpoint NFT regression term is
\begin{equation}
\mathcal{L}_{\mathrm{NFT}}^i
=\omega_i\|\Vec{v}_{\theta,i}^{+}-\Vec{u}_t^i\|^2
+(1-\omega_i)\|\Vec{v}_{\theta,i}^{-}-\Vec{u}_t^i\|^2.
\label{eq:unchanged-nft-loss}
\end{equation}
With fields evaluated at $(\Vec{x}_t^i,t)$, the residuals satisfy
\begin{equation*}
\Vec{v}_{\theta,i}^{\pm}-\Vec{u}_t^i
=\pm\beta(\Vec{v}_\theta-\Vec{v}_{\mathrm{old},n})
-(\Vec{u}_t^i-\Vec{v}_{\mathrm{old},n}).
\end{equation*}
Expanding and collecting terms independent of $\theta$ gives
\begin{align}
\mathcal{L}_{\mathrm{NFT}}^i
={}&\beta A_i\|\Vec{v}_\theta(\Vec{x}_t^i,t)-\Vec{u}_t^i\|^2 \nonumber\\
&+\beta(\beta-A_i)
\|\Vec{v}_\theta(\Vec{x}_t^i,t)-\Vec{v}_{\mathrm{old},n}(\Vec{x}_t^i,t)\|^2+C_i,
\label{eq:nft-rvm-derivation}
\end{align}
with $C_i=(1-\beta A_i)\|\Vec{u}_t^i-\Vec{v}_{\mathrm{old},n}(\Vec{x}_t^i,t)\|^2$, which does not involve $\theta$. Hence
\begin{equation}
\lambda_i=\beta A_i,\qquad
g_i=\beta(\beta-A_i),\qquad
\Vec{v}_{\texttt{anc}}=\Vec{v}_{\mathrm{old},n}.
\label{eq:nft-rvm-coefficients}
\end{equation}
Here $\lambda_i$ takes the sign of $A_i$, and $g_i<0$ when $A_i>\beta$, which requires $\beta<1$ since $A_i\leq1$. Their sum is nevertheless constant, $\lambda_i+g_i=\beta^2>0$, so the positive-curvature condition for the minimizer in Eq.~\eqref{eq:proposal-operator} holds for every value of $A_i$.

\paragraph{AWM.}
AWM \citep{xue2025awm} estimates an endpoint likelihood ratio from differences between weighted flow matching losses and adds a velocity-space anchoring penalty. We denote its anchor by $\Vec{v}_{\texttt{anc}}$, since the configuration can use a frozen reference or an EMA field. We use a frozen-reference term with weight $\beta_{\mathrm{ref}}$ (Table~\ref{tab:prefix-protocol}) and an EMA term with weight $1$ and decay $0.99$, each with uniform time weighting. We define $\beta_{\mathrm{AWM}}=\beta_{\mathrm{ref}}+1$ and $\Vec{v}_{\texttt{anc}}=(\beta_{\mathrm{ref}}\Vec{v}_{\mathrm{ref}}+\Vec{v}_{\mathrm{ema}})/\beta_{\mathrm{AWM}}$, where $\Vec{v}_{\mathrm{ema}}$ is the EMA field. Considering  $\log\widehat p_\theta$ for the flow matching surrogate of the endpoint log-density and $\rho_i=\exp(\log\widehat p_\theta-\sg[\log\widehat p_{\mathrm{old},n}])$ for the resulting ratio, the prediction gradient of the unclipped policy term together with its anchor is proportional to
\begin{align}
{}&\rho_i A_i w_{\mathrm{pol}}(t)
\bigl(\Vec{v}_\theta(\Vec{x}_t^i,t)-\Vec{u}_t^i\bigr) \nonumber\\
&+\beta_{\mathrm{AWM}}w_{\mathrm{KL}}(t)
\bigl(\Vec{v}_\theta(\Vec{x}_t^i,t)-\Vec{v}_{\texttt{anc}}(\Vec{x}_t^i,t)\bigr).
\label{eq:awm-rvm-gradient}
\end{align}
This yields the AWM mapping in Table~\ref{tab:forward matching-views}.

\paragraph{RAM.}
RAM \citep{bergmeister2026ram} regresses toward the stop-gradient target
\begin{equation}
\sg\!\left[
\Vec{v}_{\mathrm{ref}}(\Vec{x}_t^i,t)
+r_i^{\mathrm{RAM}}\bigl(\Vec{u}_t^i-\Vec{v}_\theta(\Vec{x}_t^i,t)\bigr)
\right],
\label{eq:ram-detached-target}
\end{equation}
where $r_i^{\mathrm{RAM}}$ is the transformed reward coefficient of endpoint $i$, following the notation in RAM. Since this target depends on $\Vec{v}_\theta$, we treat it as a constant when differentiating, as the stop-gradient prescribes. Differentiating the squared regression loss at a fixed target gives a prediction gradient proportional to
\begin{align}
{}&r_i^{\mathrm{RAM}}\bigl(\Vec{v}_\theta(\Vec{x}_t^i,t)-\Vec{u}_t^i\bigr) +\bigl(\Vec{v}_\theta(\Vec{x}_t^i,t)-\Vec{v}_{\mathrm{ref}}(\Vec{x}_t^i,t)\bigr),
\label{eq:ram-rvm-gradient}
\end{align}
which gives the RAM entry of Table~\ref{tab:forward matching-views}.

\section{Reference-Assisted Proposals under the Common Regression Form}
\label{app:refmix-common-loss}

This section solves the common regression of App.~\ref{app:forward-objectives} when the endpoints are drawn from the mixed proposal $M_n$ of Eq.~\eqref{eq:proposal-mixture} instead of $P_n$. As stated in Section~\ref{sec:proposal-mixing}, the coefficient rules $\lambda_i(t,A_i)$, $g_i(t,A_i)$ and the anchor $\Vec{v}_{\texttt{anc}}$ of Eq.~\eqref{eq:anchored-forward-regression} are retained, so only the distribution of the endpoints changes. App.~\ref{app:mixed-update} gives the pointwise solution, which recovers Eq.~\eqref{eq:main-nft-proposal-update} for NFT, and separates the term through which the auxiliary proposal changes it. App.~\ref{app:proposal-properties} relates this change to $D_{\mathrm{KL}}(Q_n\|P_n)$, and App.~\ref{app:proposal-contribution} bounds its contribution to the NFT solution by $(\alpha/\beta^2)\,D_{\mathrm{KL}}(Q_n\|P_n)$, the bound used in Section~\ref{sec:proposal-analysis}.

\subsection{The common regression under the mixed proposal}
\label{app:mixed-update}

To solve the regression under $M_n$, we minimize its conditional expectation given $(\Vec{x}_t,t)$ pointwise. Since $A_i$ is computed within the prompt group by Eq.~\eqref{eq:on-policy-template}, it depends on the other endpoints in the group, so this expectation is taken over the prompt group of $N_P$ current-model and $N_Q$ auxiliary endpoints (Algorithm~\ref{alg:refmix-training}. The chosen endpoint has marginal distribution $M_n$, so we denote this expectation by $\mathbb E_{M_n}$, leaving the prompt implicit. Treating the coefficients and anchor as constants with respect to $\Vec{v}$, Eq.~\eqref{eq:rvm-common-gradient} gives the gradient
\begin{align}
\nabla_{\Vec{v}}\mathbb E_{M_n}[\mathcal{L}_{\mathrm{FM}}^i\mid \Vec{x}_t,t]
={}&2\mathbb E_{M_n}[\lambda_i+g_i\mid \Vec{x}_t,t]
(\Vec{v}-\Vec{v}_{\texttt{anc}}) \nonumber\\
&-2\mathbb E_{M_n}
[\lambda_i(\Vec{u}_t-\Vec{v}_{\texttt{anc}})\mid \Vec{x}_t,t].
\label{eq:proposal-rvm-gradient}
\end{align}
If $\mathbb E_{M_n}[\lambda_i+g_i\mid \Vec{x}_t,t]>0$, the conditional loss is strictly convex in $\Vec{v}$ and has the unique pointwise minimizer
\begin{equation}
\Vec{v}_n^\star-\Vec{v}_{\texttt{anc}}
=\frac{\mathbb E_{M_n}
[\lambda_i(\Vec{u}_t-\Vec{v}_{\texttt{anc}})\mid \Vec{x}_t,t]}
{\mathbb E_{M_n}[\lambda_i+g_i\mid \Vec{x}_t,t]}.
\label{eq:proposal-operator}
\end{equation}

For NFT, substituting the coefficients of Eq.~\eqref{eq:nft-rvm-coefficients}, for which $\lambda_i+g_i=\beta^2>0$ for every $A_i$, into Eq.~\eqref{eq:proposal-operator} gives Eq.~\eqref{eq:main-nft-proposal-update}. Setting $\alpha=0$ gives $M_n=P_n$, where every advantage is computed among current-model endpoints, and Eq.~\eqref{eq:main-nft-proposal-update} reduces to the pointwise solution of the original NFT regression.

The same solution holds for AWM and RAM under their own curvature conditions. For AWM, Eq.~\eqref{eq:proposal-operator} is the minimizer of the local quadratic with the ratio $\rho_i$ held fixed, and it requires $\mathbb E_{M_n}[\rho_iA_iw_{\mathrm{pol}}(t)+\beta_{\mathrm{AWM}}w_{\mathrm{KL}}(t)\mid \Vec{x}_t,t]>0$. For RAM, Eq.~\eqref{eq:proposal-operator} is the zero of the conditional expectation of the gradient in Eq.~\eqref{eq:ram-rvm-gradient}, and it requires $\mathbb E_{M_n}[r_i^{\mathrm{RAM}}\mid \Vec{x}_t,t]>-1$.

To see how the endpoint proposal enters Eq.~\eqref{eq:proposal-operator}, we define the forward matching field of $M_n$ as
\begin{equation}
\Vec{v}_{M_n}^{\mathrm{FM}}(\Vec{x},t)
=\mathbb E_{M_n}[\Vec{u}_t\mid \Vec{x}_t=\Vec{x},t].
\label{eq:proposal-fm-field}
\end{equation}
The forward matching field of any other endpoint distribution, such as $\Vec{v}_{P_n}^{\mathrm{FM}}$, is defined in the same way. Splitting the conditional mean of $\lambda_i\Vec{u}_t$ as
\begin{equation*}
\begin{aligned}
\mathbb E_{M_n}[\lambda_i \Vec{u}_t\mid \Vec{x}_t,t]
={}&\operatorname{Cov}_{M_n}(\lambda_i,\Vec{u}_t\mid \Vec{x}_t,t)\\
&+\mathbb E_{M_n}[\lambda_i\mid \Vec{x}_t,t]\,
\Vec{v}_{M_n}^{\mathrm{FM}}
\end{aligned}
\end{equation*}
and substituting into Eq.~\eqref{eq:proposal-operator} gives
\begin{align}
\Vec{v}_n^\star-\Vec{v}_{\texttt{anc}}
={}&
\frac{\operatorname{Cov}_{M_n}(\lambda_i,\Vec{u}_t\mid \Vec{x}_t,t)}
{\mathbb E_{M_n}[\lambda_i+g_i\mid \Vec{x}_t,t]} \nonumber\\
&+
\frac{\mathbb E_{M_n}[\lambda_i\mid \Vec{x}_t,t]}
{\mathbb E_{M_n}[\lambda_i+g_i\mid \Vec{x}_t,t]}
\bigl(\Vec{v}_{M_n}^{\mathrm{FM}}-\Vec{v}_{\texttt{anc}}\bigr).
\label{eq:proposal-covariance-decomposition}
\end{align}

The first term is the conditional covariance of $\lambda_i$ and $\Vec{u}_t$, and the second scales the field difference $\Vec{v}_{M_n}^{\mathrm{FM}}-\Vec{v}_{\texttt{anc}}$ by the ratio of conditional coefficient means. For NFT, the ratio is $\mathbb E_{M_n}[A_i\mid \Vec{x}_t,t]/\beta$ and the field difference is the conditional mean of the residual $\Vec{u}_t-\Vec{v}_{\mathrm{old},n}$, so these two factors correspond to the advantage weight and the residual of Section~\ref{sec:proposal-analysis}, averaged over the endpoints that reach $\Vec{x}_t$. When auxiliary endpoints receive consistently lower rewards than current-model endpoints, most of them receive negative advantages . At noisy states that come mostly from these endpoints, the ratio can then become negative, and the second term points away from the velocity targets these endpoints supply. As discussed in Section~\ref{sec:proposal-analysis}, comparable rewards can exclude this case. Another requirement, proximity to $P_n$, concerns the field difference. 

\subsection{Endpoint proposals and forward matching fields}
\label{app:proposal-properties}

The field difference in Eq.~\eqref{eq:proposal-covariance-decomposition} splits as
\begin{equation}
\Vec{v}_{M_n}^{\mathrm{FM}}-\Vec{v}_{\texttt{anc}}
=\bigl(\Vec{v}_{P_n}^{\mathrm{FM}}-\Vec{v}_{\texttt{anc}}\bigr)
+\bigl(\Vec{v}_{M_n}^{\mathrm{FM}}-\Vec{v}_{P_n}^{\mathrm{FM}}\bigr).
\label{eq:mixture-anchor-discrepancy}
\end{equation}
The first part is present under the current-model proposal $P_n$ and does not depend on $Q_n$. The second is introduced by the auxiliary proposal, and we relate it to the KL divergence between endpoint distributions.

The relation to KL holds on the linear Gaussian path, the case $a_t=1-t$ and $b_t=t$ of Eq.~\eqref{eq:generic-analytic-noising}:
\begin{equation}
\Vec{x}_t=(1-t)\Vec{x}_0+t\Vec{\epsilon},\qquad
\Vec{u}_t=\Vec{\epsilon}-\Vec{x}_0,\qquad
\Vec{\epsilon}\sim\mathcal N(0,I).
\label{eq:theory-linear-path}
\end{equation}
For endpoint distributions $M$ and $P$, let $M_t$ and $P_t$ denote the distributions of $\Vec{x}_t$ under this path, and define
\begin{equation}
\mathcal E(M,P)
=\int_0^1\frac{1-t}{t}\,
\mathbb E_{M_t}
\left[
\|\Vec{v}_M^{\mathrm{FM}}(\Vec{x},t)-\Vec{v}_P^{\mathrm{FM}}(\Vec{x},t)\|^2
\right]dt.
\label{eq:integrated-field-discrepancy}
\end{equation}
The weight $(1-t)/t$ comes from the relation between score and velocity differences on this path.

\begin{proposition}[Endpoint KL and forward matching fields]
\label{prop:kl-field-discrepancy}
Suppose $M$ and $P$ have smooth positive densities, finite second moments, finite $D_{\mathrm{KL}}(M\|P)$, and sufficient decay for integration by parts. Under Eq.~\eqref{eq:theory-linear-path},
\begin{equation}
\mathcal E(M,P)=D_{\mathrm{KL}}(M\|P).
\label{eq:kl-field-identity}
\end{equation}
For the mixture $M_n$ of Eq.~\eqref{eq:proposal-mixture}, if $M_n$ and $P_n$ satisfy these conditions and $D_{\mathrm{KL}}(Q_n\|P_n)<\infty$, then
\begin{equation}
\mathcal E(M_n,P_n)
\leq\alpha D_{\mathrm{KL}}(Q_n\|P_n).
\label{eq:mixture-kl-discrepancy}
\end{equation}
Moreover, for fixed $P_n$ and $Q_n$, if the identity holds for every $\alpha\in[0,1]$, then $\mathcal E(M_n,P_n)$ is convex and nondecreasing in $\alpha$.
\end{proposition}

\begin{proof}
Let $m_t$ and $p_t$ denote the densities of $M_t$ and $P_t$. Given $\Vec{x}_0$, the noised state is Gaussian with mean $(1-t)\Vec{x}_0$ and covariance $t^2I$, so
\begin{equation}
\nabla\log m_t(\Vec{x})
=-\frac{\mathbb E_M[\Vec{\epsilon}\mid \Vec{x}_t=\Vec{x}]}{t}
=-\frac{\Vec{x}+(1-t)\Vec{v}_M^{\mathrm{FM}}(\Vec{x},t)}{t},
\label{eq:score-velocity-identity}
\end{equation}
where the second equality uses $\Vec{x}=\mathbb E_M[\Vec{x}_0\mid \Vec{x}_t=\Vec{x}]+t\Vec{v}_M^{\mathrm{FM}}(\Vec{x},t)$. The same identity holds for $p_t$ and $P$, and the common term $\Vec{x}/t$ cancels in the difference:
\begin{equation}
\nabla\log\frac{m_t}{p_t}
=-\frac{1-t}{t}
\bigl(\Vec{v}_M^{\mathrm{FM}}-\Vec{v}_P^{\mathrm{FM}}\bigr).
\label{eq:score-difference}
\end{equation}
The two densities satisfy the continuity equations $\partial_tm_t=-\nabla\!\cdot(m_t\Vec{v}_M^{\mathrm{FM}})$ and $\partial_tp_t=-\nabla\!\cdot(p_t\Vec{v}_P^{\mathrm{FM}})$. Differentiating the KL and using $\int\partial_tm_t=0$ gives
\begin{equation*}
\frac{d}{dt}D_{\mathrm{KL}}(M_t\|P_t)
=\int\partial_tm_t\log\frac{m_t}{p_t}
-\int\frac{m_t}{p_t}\partial_tp_t.
\end{equation*}
Integration by parts turns the first term into $\int m_t\Vec{v}_M^{\mathrm{FM}}\cdot\nabla\log(m_t/p_t)$ and, with $\nabla(m_t/p_t)=(m_t/p_t)\nabla\log(m_t/p_t)$, the second into $-\int m_t\Vec{v}_P^{\mathrm{FM}}\cdot\nabla\log(m_t/p_t)$. Adding the two and applying Eq.~\eqref{eq:score-difference} gives
\begin{equation}
\frac{d}{dt}D_{\mathrm{KL}}(M_t\|P_t)
=-\frac{1-t}{t}\,
\mathbb E_{M_t}
\|\Vec{v}_M^{\mathrm{FM}}-\Vec{v}_P^{\mathrm{FM}}\|^2.
\label{eq:relative-entropy-dissipation}
\end{equation}

The right side is nonpositive, so $D_{\mathrm{KL}}(M_t\|P_t)$ is nonincreasing in $t$ and has limits as $t\to0$ and $t\to1$. Integrating Eq.~\eqref{eq:relative-entropy-dissipation} over closed subintervals that approach $(0,1)$ gives
\begin{equation*}
\lim_{t\to1}D_{\mathrm{KL}}(M_t\|P_t)-\lim_{t\to0}D_{\mathrm{KL}}(M_t\|P_t)
=-\mathcal E(M,P),
\end{equation*}
so Eq.~\eqref{eq:kl-field-identity} follows once the limit as $t\to1$ is shown to be $0$ and the limit as $t\to0$ to be $D_{\mathrm{KL}}(M\|P)$. Since KL is not continuous under weak convergence, we verify both directly. As $t\to1$, choose a coupling $(X,Y)$ of $M$ and $P$ with finite $\mathbb E\|X-Y\|^2$; joint convexity of KL, applied to the Gaussian noising kernels, gives
\begin{equation*}
D_{\mathrm{KL}}(M_t\|P_t)
\leq\frac{(1-t)^2}{2t^2}\mathbb E\|X-Y\|^2,
\end{equation*}
which vanishes as $t\to1$. As $t\to0$, the noising kernel cannot increase KL, so $D_{\mathrm{KL}}(M_t\|P_t)\leq D_{\mathrm{KL}}(M\|P)$ for every $t$, while $M_t\to M$ and $P_t\to P$ weakly and lower semicontinuity of KL give the reverse inequality in the limit.

For the mixture, convexity of $D_{\mathrm{KL}}(\cdot\|P_n)$ in its first argument gives
\begin{align*}
D_{\mathrm{KL}}(M_n\|P_n)
&\leq(1-\alpha)D_{\mathrm{KL}}(P_n\|P_n)
+\alpha D_{\mathrm{KL}}(Q_n\|P_n)\\
&=\alpha D_{\mathrm{KL}}(Q_n\|P_n),
\end{align*}
which, with Eq.~\eqref{eq:kl-field-identity}, gives Eq.~\eqref{eq:mixture-kl-discrepancy}. The same convexity makes $D_{\mathrm{KL}}(M_n\|P_n)$ convex in $\alpha$. Since it is nonnegative and zero at $\alpha=0$, it is nondecreasing on $[0,1]$, and Eq.~\eqref{eq:kl-field-identity} transfers both properties to $\mathcal E(M_n,P_n)$.
\end{proof}

\subsection{Contribution to the regression solution}
\label{app:proposal-contribution}

Substituting Eq.~\eqref{eq:mixture-anchor-discrepancy} into Eq.~\eqref{eq:proposal-covariance-decomposition} separates the component added by the auxiliary proposal, as we define
\begin{equation}
\begin{aligned}
\Vec{\Delta}_n^{\mathrm{prop}}(\Vec{x},t)
:={}&
\frac{\mathbb E_{M_n}[\lambda_i\mid \Vec{x}_t=\Vec{x},t]}
{\mathbb E_{M_n}[\lambda_i+g_i\mid \Vec{x}_t=\Vec{x},t]}\\
&\cdot\bigl(
\Vec{v}_{M_n}^{\mathrm{FM}}(\Vec{x},t)-\Vec{v}_{P_n}^{\mathrm{FM}}(\Vec{x},t)
\bigr).
\end{aligned}
\label{eq:proposal-component}
\end{equation}
After this substitution, the solution in Eq.~\eqref{eq:proposal-covariance-decomposition} has three terms: the covariance term, the anchor part $\Vec{v}_{P_n}^{\mathrm{FM}}-\Vec{v}_{\texttt{anc}}$ scaled by the coefficient ratio, and $\Vec{\Delta}_n^{\mathrm{prop}}$. The first two are present at $\alpha=0$ and change with the proposal only through the conditional expectations under $M_n$, whereas $\Vec{\Delta}_n^{\mathrm{prop}}$ vanishes at $\alpha=0$. We bound only $\Vec{\Delta}_n^{\mathrm{prop}}$: Proposition~\ref{prop:kl-field-discrepancy} bounds the time-weighted squared norm of its field difference, $\mathcal E(M_n,P_n)$, by $\alpha D_{\mathrm{KL}}(Q_n\|P_n)$, and the coefficient ratio determines how much of this difference enters the solution.

For nonnegative $\lambda_i,g_i$ with positive conditional curvature, the ratio lies in $[0,1]$, so Eq.~\eqref{eq:integrated-field-discrepancy} and Eq.~\eqref{eq:mixture-kl-discrepancy} give
\begin{align}
\int_0^1\frac{1-t}{t}\,
\mathbb E_{M_{n,t}}
\|\Vec{\Delta}_n^{\mathrm{prop}}(\Vec{x},t)\|^2dt
&\leq\mathcal E(M_n,P_n)\nonumber\\
&\leq\alpha D_{\mathrm{KL}}(Q_n\|P_n).
\label{eq:proposal-energy-bound}
\end{align}
The NFT coefficients are signed, but the ratio equals $\mathbb E_{M_n}[A_i\mid \Vec{x}_t,t]/\beta$ (App.~\ref{app:mixed-update}), whose magnitude is at most $1/\beta$ since $A_i\in[-1,1]$. The same two equations then give
\begin{align}
\int_0^1\frac{1-t}{t}\,
\mathbb E_{M_{n,t}}
\|\Vec{\Delta}_n^{\mathrm{prop}}(\Vec{x},t)\|^2dt
&\leq\frac{\mathcal E(M_n,P_n)}{\beta^2}\nonumber\\
&\leq\frac{\alpha}{\beta^2}
D_{\mathrm{KL}}(Q_n\|P_n).
\label{eq:nft-proposal-energy-bound}
\end{align}
This is the bound stated in Section~\ref{sec:proposal-analysis}, and it formalizes a requirement, an auxiliary proposal close to $P_n$.

\section{Candidate Exposure and Proposal Discrepancy}
\label{app:exposure-prefix}

This section analyzes two properties of the auxiliary proposal: how much it increases the exposure of an underrepresented high-reward region, and how far the hybrid proposal departs from $P_n$. App.~\ref{app:discovery-discrepancy} computes the exposure and the condition under which it increases with $\alpha$, as stated in Section~\ref{sec:proposal-analysis}. App.~\ref{app:shared-suffix} shows that $D_{\mathrm{KL}}(H_n^{(k)}\|P_n)$ cannot exceed the KL divergence between the hybrid and ordinary states after the reference prefix, as stated in Section~\ref{sec:hybrid-proposal}, and App.~\ref{app:single-euler} evaluates this bound for a single reference Euler step.

\subsection{Candidate exposure}
\label{app:discovery-discrepancy}

For a prompt $\Vec{c}$, let $S$ be a distinct high-reward region that is underrepresented in $P_n$, with $p=p_n(S\mid \Vec{c})$ as in Section~\ref{sec:online-sampling} and $q=Q_n(S\mid \Vec{c})$. As in App.~\ref{app:diagnostic-interventions}, the exposure of $S$ is the probability that at least one endpoint of the pool lies in $S$ (Eq.~\eqref{eq:on-policy-exposure}). For a pool of $N_P=(1-\alpha)N$ current-model and $N_Q=\alpha N$ auxiliary endpoints drawn independently, the exposure is
\begin{equation}
1-(1-p)^{N_P}(1-q)^{N_Q}.
\label{eq:stratified-discovery}
\end{equation}
Replacing a current-model draw with an auxiliary draw multiplies the probability of missing $S$ by $(1-q)/(1-p)$. Exposure therefore increases with $\alpha$ when $q>p$, which makes precise the condition in Section~\ref{sec:proposal-analysis} that $Q_n$ assign sufficient probability mass to the region. Since every replacement multiplies this probability by the same factor, exposure is concave in $\alpha$ for $q<1$, so when $q>p$ successive replacements give diminishing gains. For RefMix, $Q_n=H_n^{(k)}$, so $q$ depends on both the reference prefix and the current-model suffix.

By contrast, the proposal discrepancy $\mathcal E(M_n,P_n)$ is convex and nondecreasing in $\alpha$ under the conditions of Proposition~\ref{prop:kl-field-discrepancy}. Raising $\alpha$ thus adds exposure with diminishing returns while the proposal discrepancy grows at a nondecreasing rate, so the two favor opposite values of $\alpha$. Section~\ref{sec:ablations} selects $\alpha$ empirically under this trade-off (Figure~\ref{fig:proposal-ablation}).

Eq.~\eqref{eq:stratified-discovery} assumes independent draws, whereas the offline bank draws its $N_Q$ auxiliary endpoints without replacement (App.~\ref{app:offline-bank}). Let $B_{\mathrm{el}}$ be the number of bank endpoints eligible for prompt $\Vec{c}$ and let $b_S$ be the number of them in $S$. Since the bank draws are independent of the current-model draws, the exposure is
\begin{equation*}
1-(1-p)^{N_P}\frac{\binom{B_{\mathrm{el}}-b_S}{N_Q}}{\binom{B_{\mathrm{el}}}{N_Q}}.
\end{equation*}

\subsection{Shared current-model suffix}
\label{app:shared-suffix}

Hybrid and ordinary rollouts use different fields only in the first $K_{\mathrm{ref}}=kK$ solver steps (Eq.~\eqref{eq:hybrid-field-general}). Let $P_n^{\mathrm{sw}}$ and $H_n^{\mathrm{sw}}$ denote their state distributions after these steps, at which the hybrid sampler switches to $\Vec{v}_{\mathrm{old},n}$. Both samplers then apply the same current-model suffix. Since all rollout samplers are deterministic, this suffix is a map $F_n$ from the switch state to the endpoint, and
\begin{equation}
P_n=(F_n)_\#P_n^{\mathrm{sw}},\qquad
H_n^{(k)}=(F_n)_\#H_n^{\mathrm{sw}},
\label{eq:shared-suffix-pushforward}
\end{equation}
where $(F_n)_\#$ maps the distribution of a random variable $X$ to that of $F_n(X)$.

\begin{proposition}[KL under a shared current-model suffix]
\label{prop:shared-suffix}
For a common measurable map $F_n$ and finite $D_{\mathrm{KL}}(H_n^{\mathrm{sw}}\|P_n^{\mathrm{sw}})$,
\begin{equation}
D_{\mathrm{KL}}(H_n^{(k)}\|P_n)
\leq D_{\mathrm{KL}}(H_n^{\mathrm{sw}}\|P_n^{\mathrm{sw}}).
\label{eq:shared-suffix-kl}
\end{equation}
\end{proposition}

\begin{proof}
Let $\zeta=dH_n^{\mathrm{sw}}/dP_n^{\mathrm{sw}}$. After applying $F_n$, the likelihood ratio becomes $\mathbb E_{P_n^{\mathrm{sw}}}[\zeta\mid F_n(X)]$ for $X\sim P_n^{\mathrm{sw}}$, and conditional Jensen applied to $s\log s$ gives
\begin{align*}
D_{\mathrm{KL}}((F_n)_\#H_n^{\mathrm{sw}}\|(F_n)_\#P_n^{\mathrm{sw}})
&\leq\mathbb E_{P_n^{\mathrm{sw}}}[\zeta\log\zeta]\\
&=D_{\mathrm{KL}}(H_n^{\mathrm{sw}}\|P_n^{\mathrm{sw}}).
\end{align*}
\end{proof}

\begin{figure*}[!t]
\centering
\begin{minipage}[c]{0.60\textwidth}
\centering
\small
\setlength{\tabcolsep}{4pt}
\begin{tabular}{lccccc}
\toprule
Method & PickScore$\uparrow$ & $D_{\mathrm{all}}\uparrow$ & $D_{\mathrm{LPIPS}}\uparrow$ & CLIPScore$\uparrow$ & VideoAlign$\uparrow$ \\
\midrule
Reference      & 18.72 & 0.202 & 0.579 & 0.211 & $-3.20$ \\
NFT            & 21.26 & \textbf{0.086} & 0.423 & 0.253 & $-0.90$ \\
NFT + RefMix   & \textbf{21.47} & \textbf{0.086} & \textbf{0.499} & \textbf{0.269} & $\mathbf{-0.61}$ \\
\bottomrule
\end{tabular}
\end{minipage}%
\hfill
\begin{minipage}[c]{0.36\textwidth}
\centering
\includegraphics[width=0.8\linewidth]{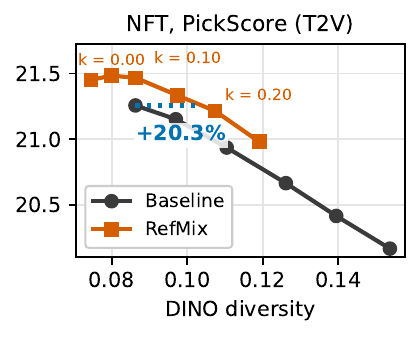}
\end{minipage}
\vspace{-0.3cm}
\caption{Text-to-video results on Wan2.1-T2V-1.3B with the PickScore reward. Left: NFT uses standard inference and NFT + RefMix uses hybrid inference with $k_{\mathrm{test}}=0.10$. Right: inference intervention at the fixed checkpoints, varying $k_{\mathrm{test}}$.}
\label{fig:t2v}
\end{figure*}

\subsection{A single reference Euler step}
\label{app:single-euler}

To evaluate the right side of Eq.~\eqref{eq:shared-suffix-kl} in closed form, assume that $\Vec{v}_{\mathrm{ref}}$ and $\Vec{v}_{\mathrm{old},n}$ are the exact forward matching fields, under Eq.~\eqref{eq:theory-linear-path}, of distributions $P_{\mathrm{ref}}$ and $P_{\mathrm{tgt}}$ with finite second moments and means $\Vec{\mu}_{\mathrm{ref}}$ and $\Vec{\mu}_{\mathrm{tgt}}$, that is, $\Vec{v}_{\mathrm{ref}}=\Vec{v}_{P_{\mathrm{ref}}}^{\mathrm{FM}}$ and $\Vec{v}_{\mathrm{old},n}=\Vec{v}_{P_{\mathrm{tgt}}}^{\mathrm{FM}}$. The endpoint distribution $P_n$ of the finite-step sampler can still differ from $P_{\mathrm{tgt}}$ because of discretization.

Use $K\geq2$ uniform explicit Euler steps from $t=1$ to $t=0$ with step size $1/K$, and apply the reference field only at the first step, so that $K_{\mathrm{ref}}=1$ and $k=1/K$. At $t=1$, the noised state $\Vec{x}_1=\Vec{\epsilon}$ is independent of $\Vec{x}_0$, so at a sampler state $\Vec{z}$,
\begin{gather*}
\Vec{v}_{P_{\mathrm{tgt}}}^{\mathrm{FM}}(\Vec{z},1)=\Vec{z}-\Vec{\mu}_{\mathrm{tgt}},\\
\Vec{z}-\frac1K \Vec{v}_{P_{\mathrm{tgt}}}^{\mathrm{FM}}(\Vec{z},1)
=\left(1-\frac1K\right)\Vec{z}+\frac{\Vec{\mu}_{\mathrm{tgt}}}{K},
\end{gather*}
and likewise for $P_{\mathrm{ref}}$ with $\Vec{\mu}_{\mathrm{ref}}$. Since the initial state is $\Vec{z}\sim\mathcal N(0,I)$, the switch distributions are Gaussian even when $P_{\mathrm{ref}}$ and $P_{\mathrm{tgt}}$ are not:
\begin{align}
P_n^{\mathrm{sw}}&=\mathcal N\!\left(
\frac{\Vec{\mu}_{\mathrm{tgt}}}{K},(1-1/K)^2I
\right), \nonumber\\
H_n^{\mathrm{sw}}&=\mathcal N\!\left(
\frac{\Vec{\mu}_{\mathrm{ref}}}{K},(1-1/K)^2I
\right).
\label{eq:gmm-switch-laws}
\end{align}
Their Gaussian KL and Eq.~\eqref{eq:shared-suffix-kl}, together with Eq.~\eqref{eq:mixture-kl-discrepancy}, give
\begin{align}
D_{\mathrm{KL}}(H_n^{(1/K)}\|P_n)
&\leq D_{\mathrm{KL}}(H_n^{\mathrm{sw}}\|P_n^{\mathrm{sw}})\nonumber\\
&=\frac{\|\Vec{\mu}_{\mathrm{ref}}-\Vec{\mu}_{\mathrm{tgt}}\|^2}
{2(K-1)^2},
\label{eq:one-step-hybrid-kl}\\
\mathcal E(M_n,P_n)
&\leq
\frac{\alpha\|\Vec{\mu}_{\mathrm{ref}}-\Vec{\mu}_{\mathrm{tgt}}\|^2}
{2(K-1)^2},
\label{eq:one-step-mixture-kl}
\end{align}
where the last line also requires the conditions of Proposition~\ref{prop:kl-field-discrepancy}.

In this idealization, the reference step shifts every state by the same vector $(\Vec{\mu}_{\mathrm{ref}}-\Vec{\mu}_{\mathrm{tgt}})/K$ relative to the current-model step from the same noise. Only the means of $P_{\mathrm{ref}}$ and $P_{\mathrm{tgt}}$ therefore enter the switch distributions, and equal means give identical prefixes even if the two distributions differ in diversity. Since the shared suffix $F_n$ is nonlinear, this uniform shift can still change which region an endpoint reaches, which is the role of the reference prefix in Section~\ref{sec:hybrid-proposal}. At the same time, Eq.~\eqref{eq:one-step-hybrid-kl} bounds the KL divergence between the resulting endpoint distributions.

The one-step case matches the prefix length of the NFT and AWM training rollouts, one reference step out of $K=10$ and $K=14$, respectively. It idealizes these rollouts with exact fields and Euler steps, whereas NFT uses DPM2 and AWM a second-order SA-Solver. With more than one reference step, as in RAM ($K_{\mathrm{ref}}=2$ of $K=20$), later field evaluations depend on intermediate states and the switch distributions are in general no longer Gaussian, so this calculation does not extend to that case.

\section{Extension to Text-to-Video Generation}
\label{app:t2v}

RefMix only changes where training endpoints come from and makes no assumption about the data modality. We therefore apply it without modification to a
text-to-video flow model. 

\paragraph{Setup.}
We fine-tune Wan2.1-T2V-1.3B using NFT and generate 17-frame $480\times832$ videos. The reward is PickScore averaged over all frames. Each update uses 24 prompts with $N=24$ endpoints per prompt, for 40 updates. As in the image experiments, the reference prefix is $k=0.10$ in both training and inference. We use $\alpha=0.5$. Evaluation uses 48 held-out prompts with 8 videos per prompt and computes the DINOv3 diversity per-video. As an independent video-level check, we also report the VideoAlign overall score, which is not used for training.

\begin{figure*}[t]
\centering
\includegraphics[width=\textwidth]{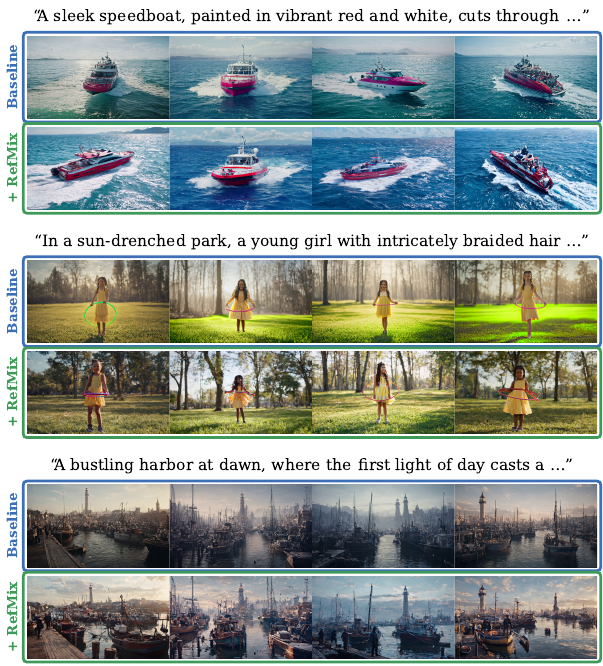}
\caption{Qualitative text-to-video comparison on Wan2.1-T2V-1.3B with the PickScore reward. Each panel shows the middle frame of videos generated from four initial noises. Baseline videos converge to a similar viewpoint and appearance and develop oversaturated backgrounds, whereas RefMix retains alternatives in viewpoint, framing, and scene.}
\label{fig:t2v_qual}
\end{figure*}

\paragraph{Results.}
Figure~\ref{fig:t2v} (left) shows the same pattern as on images: RefMix reaches a higher reward than the baseline while keeping its visual diversity, and its videos differ more at the pixel level ($D_{\mathrm{LPIPS}}$). RefMix also raises CLIPScore and the held-out VideoAlign score, so the higher PickScore does not come from exploiting the training reward alone. Figure~\ref{fig:t2v} (right) separates the training and inference effects as in Sec.~\ref{sec:training-inference}. The RefMix-trained model lies above that curve, so at the reward of standard baseline inference it provides 20.3\% more DINO diversity  comparable to that on same reward scale. Figure~\ref{fig:t2v_qual} shows qualitative results. The baseline repeats a similar composition, approaching the boat from nearly the same angle. RefMix varies the viewpoint and framing and keeps a natural appearance while preserving the prompt content.

\section{Additional Qualitative Comparisons}
\label{app:qualitative}
\begin{figure*}[p]
\centering
\includegraphics[width=\textwidth]{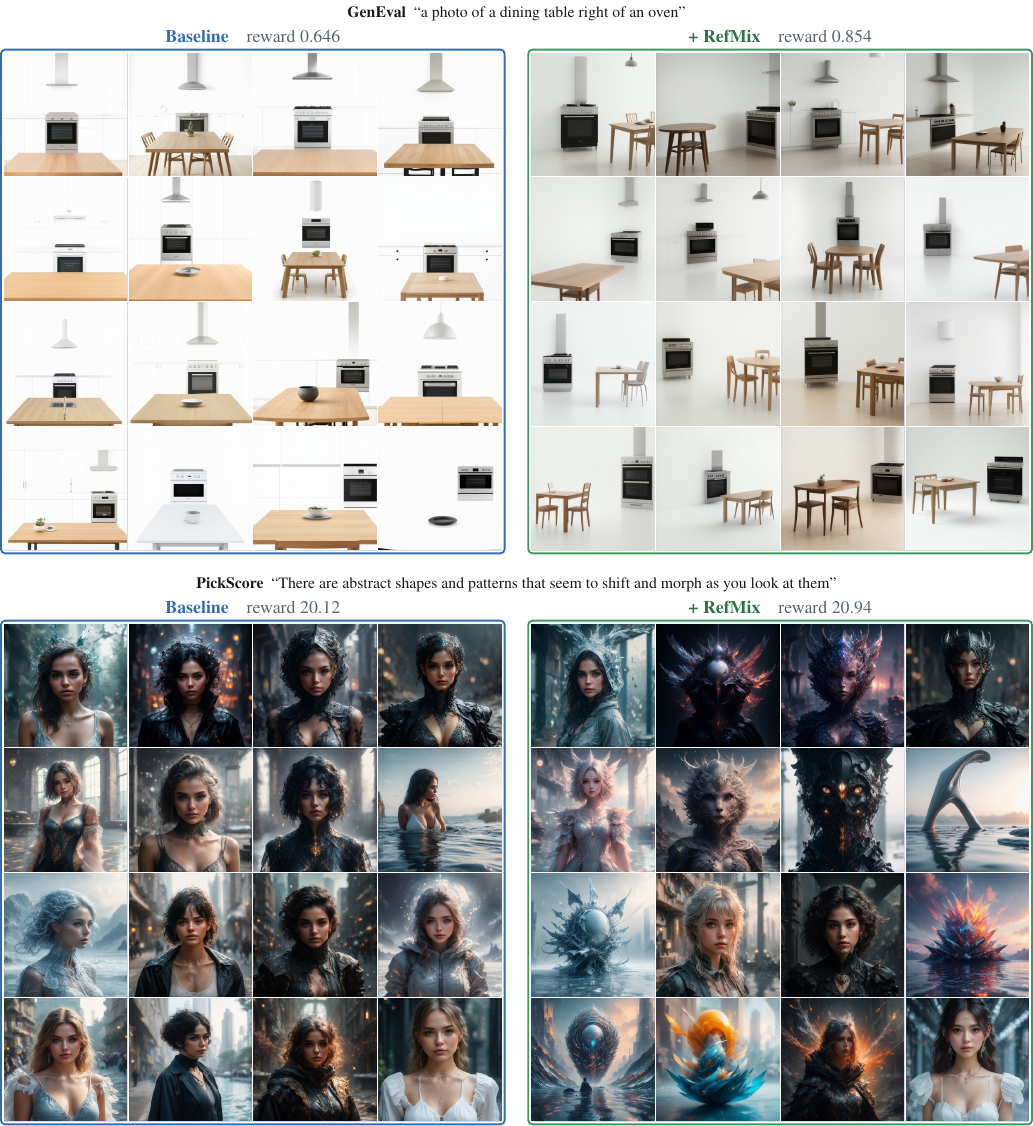}
\caption{Additional NFT samples. All 16 generations for a GenEval prompt (top) and a PickScore prompt (bottom), from the baseline (left) and from RefMix with four reference calls out of 40 at inference (right). Tiles at the same position share the initial noise. Each label gives the mean target reward of the 16 samples, in the units of Table~\ref{tab:main-results}.}
\label{fig:app-qual-nft}
\end{figure*}

\begin{figure*}[p]
\centering
\includegraphics[width=\textwidth]{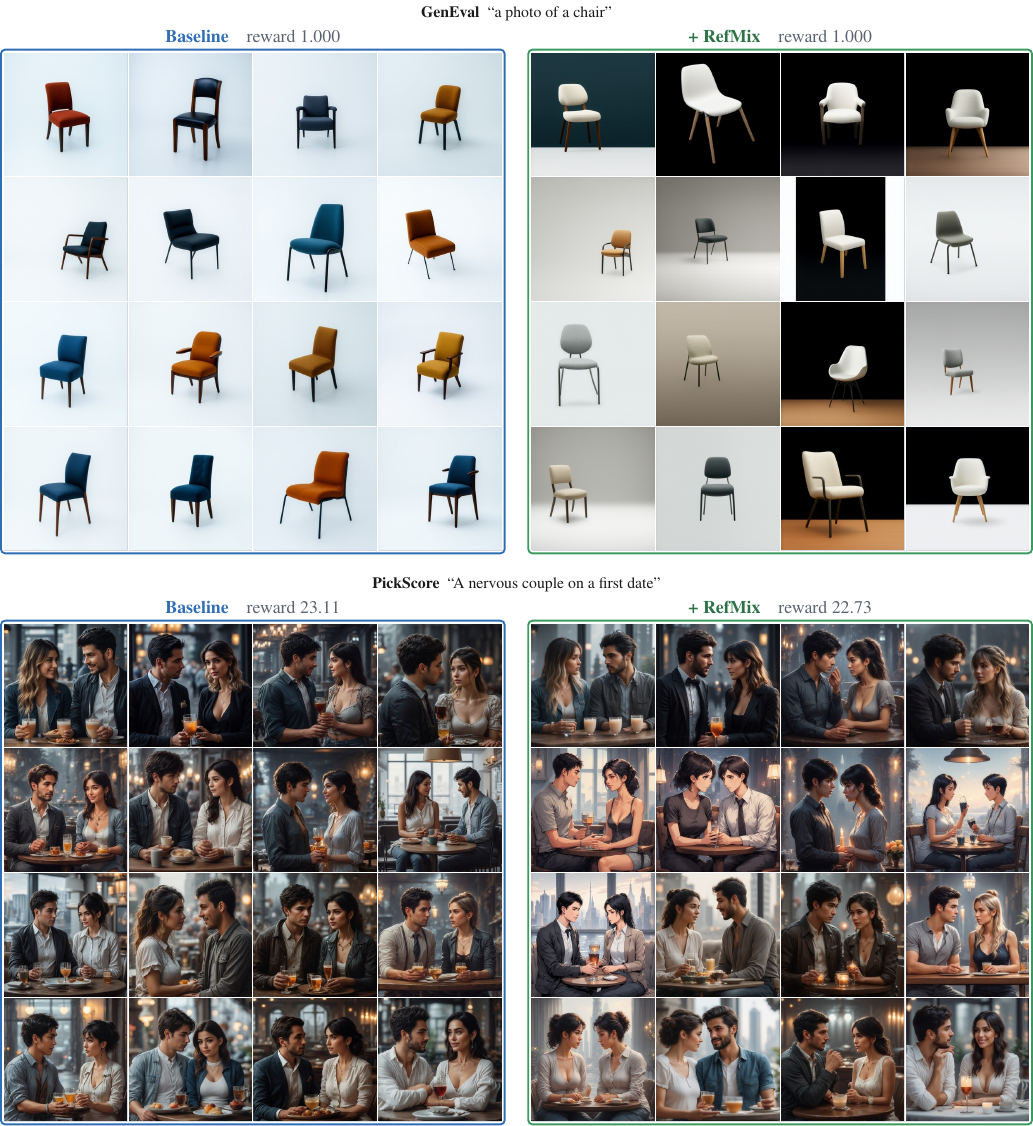}
\caption{Additional AWM samples. All 16 generations for a GenEval prompt (top) and a PickScore prompt (bottom), from the baseline (left) and from RefMix with three reference calls out of 40 at inference (right). Tiles at the same position share the initial noise. Each label gives the mean target reward of the 16 samples, in the units of Table~\ref{tab:main-results}.}
\label{fig:app-qual-awm}
\end{figure*}

\begin{figure*}[p]
\centering
\includegraphics[width=\textwidth]{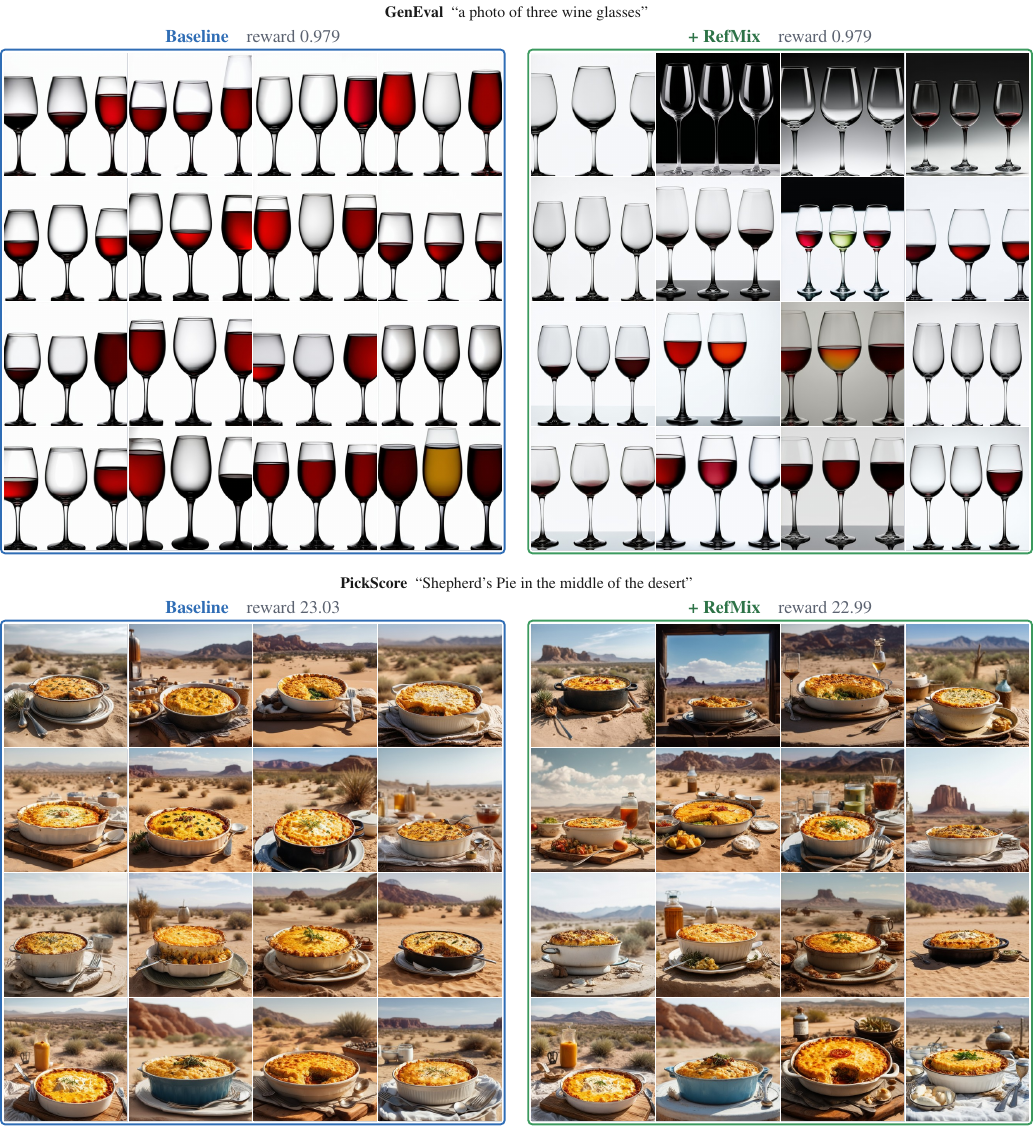}
\caption{Additional RAM samples. All 16 generations for a GenEval prompt (top) and a PickScore prompt (bottom), from the baseline (left) and from RefMix with four reference calls out of 40 at inference (right). Tiles at the same position share the initial noise. Each label gives the mean target reward of the 16 samples, in the units of Table~\ref{tab:main-results}.}
\label{fig:app-qual-ram}
\end{figure*}

Figures~\ref{fig:app-qual-nft}--\ref{fig:app-qual-ram} show all $C=16$ generations for GenEval prompts and PickScore prompts per method. Checkpoints, guidance, and inference follow the evaluation protocol of Appendix~\ref{app:experimental-details}: the baseline uses current-model-only inference and RefMix uses hybrid inference with the test prefix of Table~\ref{tab:prefix-protocol}. Both arms share the prompt and the initial noise. 

\end{document}